\documentclass[sigconf, anonymous=false,10pt]{acmart}
\renewcommand\footnotetextcopyrightpermission[1]{}
\usepackage{algpseudocode} 
\usepackage{algorithm}
\usepackage{subcaption}
\usepackage{listings}
\usepackage{booktabs} 
\usepackage{alltt}
\definecolor{bluekeywords}{rgb}{0.13,0.13,1}
\definecolor{greencomments}{rgb}{0,0.5,0}
\definecolor{greynumber}{rgb}{0.6,0.6,0.6}
\definecolor{redstrings}{rgb}{0.9,0,0}
\newcommand{\head}[1]{\noindent \textbf{#1}}

\setcopyright{rightsretained}

\acmDOI{10.475/123_4}

\acmISBN{123-4567-24-567/08/06}

\acmArticle{4}
\acmPrice{15.00}

\begin{document}
\title{LIFT: Reinforcement Learning in Computer Systems by Learning From Demonstrations}

\author{Michael Schaarschmidt}
\affiliation{\institution{University of Cambridge}}
\email{michael.schaarschmidt@cl.cam.ac.uk}

\author{Alexander Kuhnle}
\affiliation{\institution{University of Cambridge}}
\email{alexander.kuhnle@cl.cam.ac.uk}

\author{Ben Ellis}
\affiliation{\institution{University of Cambridge}}
\email{be255@cam.ac.uk}

\author{Kai  Fricke}
\affiliation{\institution{Helmut Schmidt University}}
\email{fricke@hsu-hh.de}

\author{Felix Gessert}
\affiliation{\institution{Baqend}}
\email{fg@baqend.com}

\author{Eiko Yoneki}
\affiliation{\institution{University of Cambridge}}
\email{eiko.yoneki@cl.cam.ac.uk}

\begin{abstract}
Reinforcement learning approaches have long appealed to the data management community due to their ability to learn to control dynamic behavior from raw system performance. Recent successes in combining deep neural networks with reinforcement learning have sparked significant new interest in this domain. However, practical solutions remain elusive due to large training data requirements, algorithmic instability, and lack of standard tools.

In this work, we introduce LIFT, an end-to-end software stack for applying deep reinforcement learning to data management tasks. While prior work has frequently explored applications in simulations, LIFT centers on utilizing human expertise to learn from demonstrations, thus lowering online training times. We further introduce TensorForce, a TensorFlow library for applied deep reinforcement learning exposing a unified declarative interface to common RL algorithms, thus providing a backend to LIFT. We demonstrate the utility of LIFT in two case studies in database compound indexing and resource management in stream processing. Results show LIFT controllers initialized from demonstrations can outperform human baselines and heuristics across latency metrics and space usage by up to $70\%$.
\end{abstract}

%
%



\maketitle

\section{Introduction}
Model-free reinforcement learning (RL) techniques offer a generic framework for optimizing decision making from raw feedback signals such as system performance~\cite{SuttonBarto1998}, thus not requiring an analytical model of the system. In recent years, deep reinforcement learning (DRL) approaches which combine RL with deep neural networks have enjoyed successes in a variety of domains such as games (Go \cite{SilverHuangMaddisonEtAl2016}, Atari~\cite{MnihKavukcuogluSilverEtAl2013,double_dqn,MnihDQN2015,Espeholt2018}), and applied domains such as industrial process control \cite{Hein2017a} or robotic manipulation \cite{Tobin2017}. RL approaches have also long appealed to computer systems researchers, with experimental applications in domains such as adaptive routing or server resource management spanning back over 20 years \cite{KumarMiikkulainen1997, KumarMiikkulainen1999, TesauroJongDasEtAl2006,TesauroDasChanEtAl2007}. The advent of deep RL in in combination with widely available deep learning frameworks has renewed interest in this approach. More recent examples include automated TensorFlow device placements \cite{mirhoseini2017device, hierarchical2018}, client-side bit-rate selection for video streaming \cite{Mao2017}, and simplified cluster scheduling \cite{Mao2016}.

However, practical RL deployments in computer systems and data management remain difficult due to large training data requirements and expensive decision evaluations (e.g. multiple minutes to deploy a cluster configuration). RL algorithms also suffer from inferior predictability and stability compared to simpler heuristics \cite{Henderson2017, Mania2018}. Consequently, proof-of-concept successes in simplified and highly controlled simulations have infrequently lead to practical deployments. Nonetheless, DRL remains appealing as it combines the ability of deep neural networks to identify and combine features in unforeseen ways with learning from raw system feedback. The long-term aim is to automate manual feature and algorithm design in computer systems and potentially learn complex behaviour outperforming manual designs.

In this work, we explore these limitations by outlining a software stack for practical DRL, with focus on guiding learning via existing log data or demonstrated examples. The key idea of our paper is that in modern data processing engines, fine-granular log data can be used to extract demonstrations of desired dynamic configurations. Such demonstrations can be used to pretrain a control model, which is subsequently refined when deployed in its concrete application context. To this end, we make the following contributions:

\begin{figure*}[t] 
\centering
\includegraphics[scale=.5]{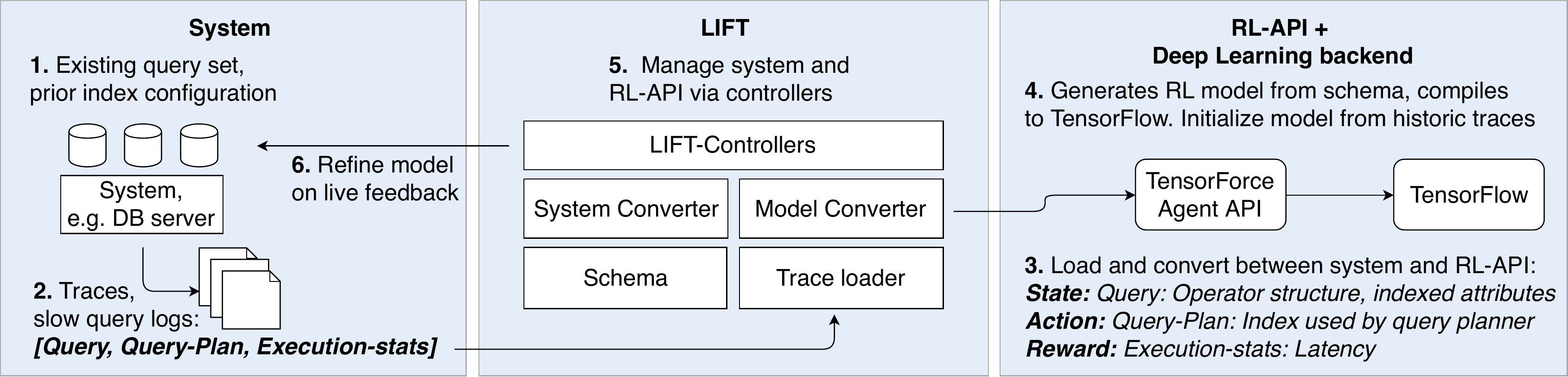}
\caption{LIFT workflow.}
\label{fig:lift-overview}
\vspace{-5mm}
\end{figure*}

We present \textit{LIFT} (\S \ref{lift}), a high level framework for \textbf{L}earn\textbf{I}ng \textbf{F}rom \textbf{T}races which provides common components to interface and map between systems and reinforcement learning, thus removing boilerplate code. We further introduce TensorForce, a highly modularized DRL library focusing on a declarative API for common algorithms, which serves as an algorithmic backend for LIFT. LIFT allows users to specify data layouts of states and action spaces which are used by TensorForce to generate TensorFlow models for executing RL tasks (\S \ref{tensorforce}). In the evaluation (\S \ref{evaluation}), we demonstrate the utility of our LIFT prototype in two experimental data management case studies. First, we use LIFT to generate a controller for automatic compound database indexing (\S \ref{traces}). Indexing is an attractive use case for RL as the optimal index set for an application depends on the complex interaction of workload, query operators within each query, data distribution, and query planner heuristics. While analytical solutions are difficult to build and vary per database and query planner, rich feedback from slow query logs enables RL controllers to identify effective solutions. Experimental results show that a LIFT-controller pretrained from imperfect rule-based demonstrations can be refined within few hours to outperform various rule and expert baselines by up to $70\%$. We also use LIFT to learn task parallelism configurations on Heron \cite{Kulkarni2015}, a state of the art stream processing engine.

Figure \ref{fig:lift-overview} illustrates LIFT's role and components. The slow query log from a database containing queries, the executed query plan, and execution statistics are read into LIFT. Via a user-defined schema and converter, LIFT interprets traces and/or provided rules as demonstrations to train an offline model. In the indexing case study, this is achieved by mapping query shape and existing indices to a state, the command required to create the index used to an action, and query performance to a reward. Traces must hence contain not only runtime performance but also corresponding configurations which can be used to reconstruct a command (action) leading to that configuration. For example, the slow query log may contain the query plan including index used, and this can be converted to the command creating that index. Schema layouts are passed to TensorForce to generate a corresponding TensorFlow graph. The states, actions, and rewards are then used to train a controller model to adopt the strategy (e.g. hand-designed rule or expert decision) behind prior indexing. Finally, LIFT is deployed in online mode to either refine indexing on an existing query set, or within a new application to replace manual tuning.

\section{Background}\label{background}
We give a brief introduction to RL with focus on practical concerns. RL is not a single optimization strategy but a class of methods used to solve the reinforcement learning problem. Informally, RL is utilized when no supervised feedback for a decision is available but reward signals indicating relative performance. For example, a cluster scheduler allocating tasks to resources may receive feedback from task completion times, but not whether a scheduling decision was optimal. 

We consider the classic formulation wherein an agent interacts with an environment $\epsilon$ described by states $s \in \mathcal{S}$ and aims to learn a policy $\pi$ that governs which action $a \in \mathcal{A}$ to take in each state \cite{SuttonBarto1998}. At each discrete time step $t$, the agent takes an action $a_t$ according to its  policy $\pi(a|s)$, transitions into a new state $s_{t+1}$ according to the environment dynamics, and observes a reward $r_t$ from a reward function $R(s,a)$. The goal of the agent is to maximize cumulative expected rewards $R = \mathbb{E}[\sum_t \gamma^{t}r_t]$, where future rewards are discounted by $\gamma$. State transitions and rewards are often assumed to be stochastic, and to satisfy the Markov property so each transition only depends on the prior state $s_{t-1}$. 

In data management tasks, the state is typically represented as a combination of the current workload and configuration, embedded into a continuous vector space. To deal with the resulting large state spaces and generalize from seen to unseen states, RL is used in conjunction with \textit{value function approximators} such as neural networks where the expected cumulative return from taking an action $a$ in state $s$  is estimated by a function parametrized by trainable parameters $\theta$ (i.e. the neural network weights). Formally, the action-value function $Q^\pi$ is given as 
\begin{align}
Q^{\pi}(s,a;\theta)=\mathbb{E}[R_t|s_t=s,a].
\end{align}
The goal of learning is to determine the optimal $Q^*(s,a)$ which maximizes expected returns. Concretely, when using Q-learning based algorithms, the neural network produces in its final layer one output per action representing it Q-value. The resulting policy is implicitly derived by greedily selecting the action with the highest Q-value while occasionally selecting random actions for exploration. Updates are performed by performing iteratively (over a sequence indexed by $i$) gradient descent on the loss~$J(\theta)_i$ \cite{MnihDQN2015}:
\begin{align}
J_i(\theta)_i = \mathbb{E}_{s,a\sim \pi}[(y_i - Q(s,a;\theta_i))^2]
\end{align}
with $y=R(s,a)+ \gamma~max_{a'}Q(s',a';\theta_{i-1})$. Intuitively, this loss is the (squared) difference between the observed reward when taking $a$ in $s$ plus the discounted estimate of future returns from the new state $s'$, and the current estimate of $Q(s,a;\theta)$, or in other words how much the Q-function has to be modified to account for observing a new reward.

In Deep Q-learning as introduced by Mnih et al. \cite{MnihDQN2015}, experience tuples of the form $(s_t,a_t,r_t,s_{t+1})$ are collected and inserted into a replay buffer, and updates are performed by sampling random batches to compute gradients. Further, learning is stabilized by using a separate target network to evaluate the Q-target $y$, which is only synchronized with the training network with delay. In contrast, policy gradient (PG) methods directly update a parametrized policy function $pi(a|s;\theta)$ such as a Gaussian or categorical distribution. This is typically (e.g. in the classical REINFORCE algorithm \cite{Williams1992}) achieved by obtaining a sample estimates of current policy performance and updating $\theta$ in the direction $\nabla_\theta log~\pi(a_t|s_t;\theta)(R_t - b_t(s_t))$. Detailed surveys of contemporary work are given by Li and Arulkumaran et al. \cite{li2017deep, arulkumaran2017brief}.

RL approaches remain attractive due to their theoretical value proposition to learn from raw feedback. However, despite over two decades of research on RL in computer systems, practical applications remain difficult to realize due to various limitations. In the following, we discuss concrete issues before introducing LIFT.



\section{Practical issues}\label{practical}
RL algorithms are known to suffer from various limitations which we highlight here in the context of data management.

\head{Training data requirements.}
First, RL methods are notoriously sample-inefficient and solving common benchmark tasks (e.g. Atari) in simulators can require up to $10^7$-$10^9$ problem interactions (states) when using recent approaches \cite{Espeholt2018}. In data management experiments, performing a single step (e.g. a scheduling decision) and observing its impact may take between seconds and hours (e.g. deciding on resources for a job and evaluating its runtime). Consequently, training  through online interaction can be impractical for some tasks, and training in production systems is further undesirable as initial behavior is random to explore. A common strategy to accelerate training is to train RL agents in simulation \cite{Mao2016, Mao2017}. This approach enables researchers to explore proof-of-concept experiments but also introduces the risk of making unrealistic assumptions and oversimplifying the problem domain, thus making successful simulation-to-real transfer unlikely. Some research domains have access to verified simulators (e.g. network protocols) but this is not the case for many ad-hoc problems in data management. 

Another common approach is to execute online training on a staging environment or a smaller deployment of the system. For example, in their recent work on hierarchical device placement in TensorFlow \cite{hierarchical2018}, Mirhoseini et al. report that training their placement mechanism on a small scale deployment for 12.5 GPU-hours saves 265 GPU hours in subsequent training of a neural network. Here, RL was used as a direct search mechanism where the aim of training is to identify a single final configuration which is not modified later. Successful online training is further difficult if the goal of the controller is to react to unpredictable and sudden workload changes. This is because training workloads may not sufficiently cover the state space to generalize to drastic workload changes (while exploring the state space is usually possible in simulation).

\head{Hyper-parameters and algorithmic stability.}
DRL algorithms require more configuration and hyper-parameter tuning than other machine learning approaches, as users need to tune neural network hyper-parameters, design of states/actions and rewards, and parameters of the reinforcement learning algorithm itself. A growing body of work in DRL attempts to address algorithmic limitations by more efficiently re-using training data, reducing variance of gradient estimates, and parallelizing training (especially in simulations) \cite{schulman2015trust, schulman2017proximal,Haarnoja2017, Espeholt2018}. Some of these efforts have recently received scrutiny as they have been shown difficult to reproduce~\cite{Henderson2017,Mania2018}, often due to the introduction of various additional hyper-parameters which again need to be tuned. This is complicated by the fact that RL algorithms are often evaluated on the task they were trained on (i.e. testing performance on the game the algorithm was trained on). RL is effectively used for optimization on a single task, and, as Mania et al. argue \cite{Mania2018}, some algorithmic improvements in recent work may stem from overfitting rather than fundamental improvements. 

\head{Software tools.}
The reproducibility issues of RL algorithms are further exacerbated by a lack of standard tools.The practical successes of neural networks in diverse domains have led to the existence of widely adopted deep learning frameworks such as Google's TensorFlow \cite{abadi2016tensorflow}, Microsoft's CNTK \cite{seide2016cntk}, or Apache MXnet \cite{chen2015mxnet}. These libraries provide common operators for implementing and executing machine learning algorithms while also omitting the complexity of directly interfacing hardware accelerators (e.g. GPUs, FPGAs, ASICs). However, RL algorithms cannot be used with similar ease as existing research code bases primarily focus on simulation environments, and thus require significant modifications to be used in practical applications. We introduce our RL library built on top of TensorFlow in section \S\ref{tensorforce}.

The issues above continue to present significant obstacles in using RL. We investigate means to improve data efficiency and tooling by providing a software stack for deep RL focused on initializing controllers from pre-existing knowledge.

\section{LIFT}\label{lift}
\subsection{System overview}
\begin{figure}[t] 
\centering
\includegraphics[scale=.4]{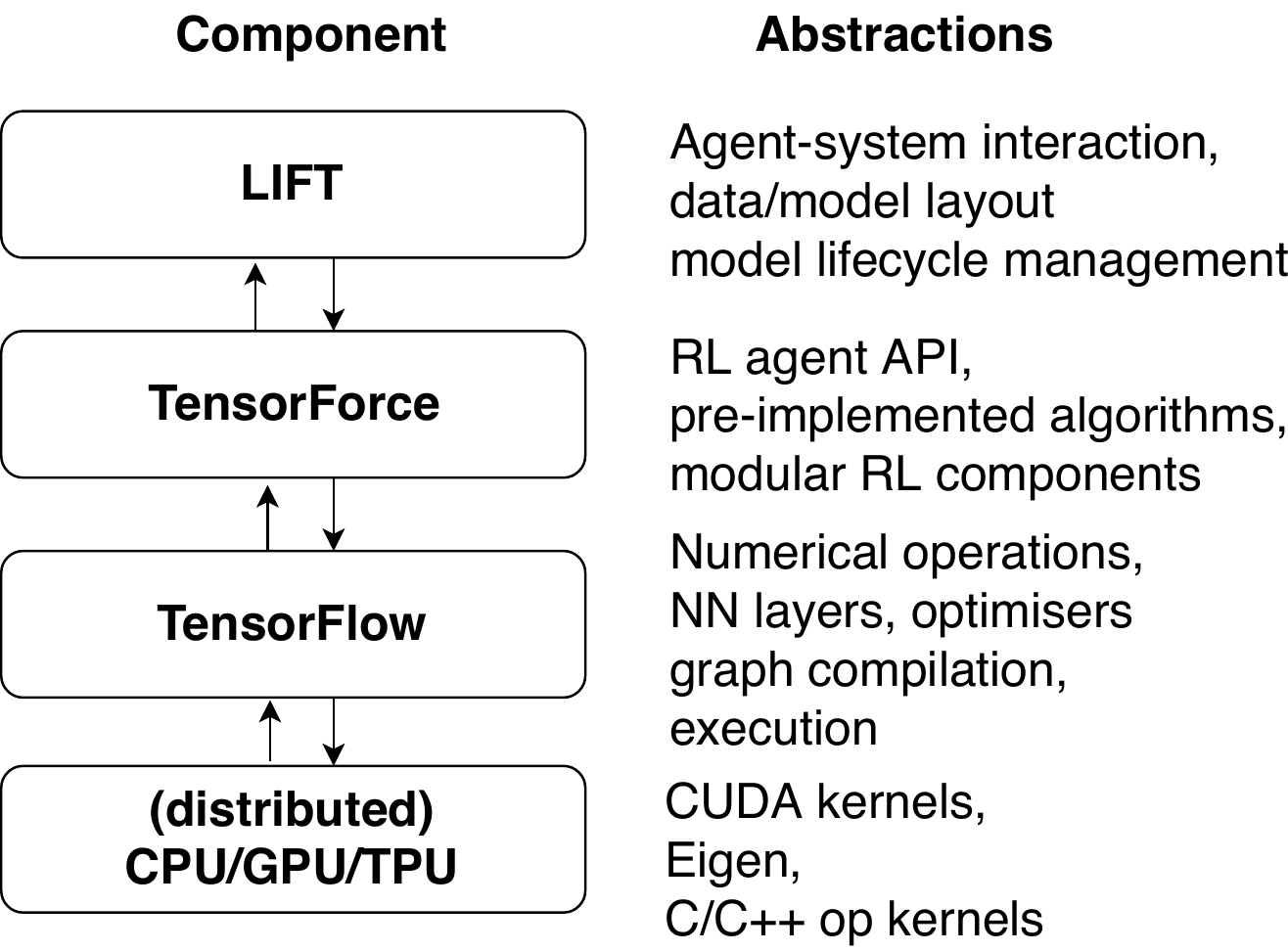}
\caption{LIFT stack for applied RL.}
\label{fig:rl-stack}
\vspace{-5mm}
\end{figure}

We begin by giving a high level overview of our framework before discussing each component in detail. Generally, we distinguish between our algorithmic backend \textit{TensorForce}, and \textit{LIFT}, a collection of services which allow RL controllers to be deployed in different execution contexts, which we explain below (Figure \ref{fig:rl-stack}). Frameworks such as TensorFlow \cite{AbadiAgarwalBarhamEtAl2016} expose an API primarily on the abstraction level of numerical operators with an increasing number of modules containing neural network layers, optimizers, probability distributions, data set tools etc. However, currently no such modules exist within TensorFlow to expose RL functionality via similar APIs. TensorForce fills this gap by providing a unified API to a set of standard RL algorithms on top of TensorFlow. 

The main abstractions LIFT operates on are RL models and system models. A model maps between RL agent output to system actions (e.g. configuration changes), or from system metrics to RL agent (e.g. parsing log entries to states, actions and rewards). LIFT's primary purpose is to facilitate RL usage in new systems by providing commonly used functionality pertaining to model serialization and evaluation, and further by defining system data layout and automatically mapping them to the respective TensorFlow inputs and outputs. LIFT uses TensorForce as its backend in our example implementation but is independent of both TensorForce and TensorFlow, as to be able to use any RL implementation providing a minimal common API. In the following, we discuss the design of TensorForce.

\subsection{TensorForce}\label{tensorforce}
Deep reinforcement learning is a rapidly evolving field and few standards exist with regard to usage outside controlled simulations. Various open source libraries such as OpenAI baselines \cite{openaibaselines}, Nervana coach \cite{nervanacoach}, or Ray Rllib \cite{Liang2017} exist. They are tightly coupled with simulation environments such as OpenAI gym \cite{brockman2016openai} which provide unified interfaces to tasks for evaluating and comparing algorithms. In our experiments, we have found these research frameworks to be difficult to deploy in practical use cases for two additional reasons.

First, open source reinforcement learning libraries frequently rely on fixed neural network architectures. For example, the code we analyzed typically created network output layers for actions based on descriptors provided by simulations only supporting restricted actions (e.g. only either discrete or continuous actions per step, but not both). Substantial code modifications are required to support multiple separate types of actions (tasks) per step. This is because the purpose of these reference implementations is primarily to reproduce research results on a particular set of benchmark tasks, as opposed to providing configurable, generic models. Second, as discussed in \S \ref{practical}, recent RL methods incorporate various optimization heuristics to help training efficiency and stability, thus increasing the number of tunable parameters. We found existing code bases to attempt reducing complexity by hard-coding heuristics of which users may be unaware. For example, one of the implementations we surveyed internally smoothes state vectors via an exponentially moving average, and clips reward values without documenting or exposing this feature. We hence introduce TensorForce, a general purpose DRL library which exposes a well-defined declarative interface to creating and transparently configuring state-of-the art algorithms.

\head{Design.}
Our aim is to give a unified interface to specify a decision model by describing its inputs and outputs without any restriction on the number and type of different inputs (states) or outputs (actions). Further, the specification contains the model to construct, network layers to use, and various further options to be applied such as exploration, input preprocessing (e.g. normalization or down-sampling) and outpost post-processing (e.g. noise), and algorithm-specific options such as memory size. 

TensorForce is built on two principles: First, users should not be required to modify any library code to express their problem dynamics, as is often the case in current open source code, thus necessitating expressive configurations. Second, reinforcement learning use cases may drastically differ in design, e.g. environments may present continuous learning or episodic problems, algorithms may use memories to incorporate old experiences, or just learn from new observations. However, most of this arising complexity can be deterministically (depending on the model selected) handled internally. Consequently, we provide a unified API for all model and agent variants with just two methods at its core, one to request new actions for given states, one to observe rewards and notify the model of terminal states. Updates to the model are implicitly triggered according to configurations.

The advantage to our approach is that practitioners can explore different RL paradigms in their applications simply by loading another configuration without the need to modify application code (e.g. to explicitly trigger certain updates or model-specific events), or library code. The code is available open source under \url{https://github.com/reinforceio/tensorforce}.

 
\head{Features.}
TensorForce implements both classical algorithms serving as an entry point for practitioners as well as newer methods, which we briefly describe. From the family of Q-learning algorithms, our library implements  the original deep Q-learning \cite{MnihDQN2015}, double deep Q-learning \cite{double_dqn}, normalized advantage functions for continuous Q-learning \cite{GuLillicrapSutskeverEtAl2016}, n-step Q-learning \cite{MnihBadiaMirzaEtAl2016}, and deep Q learning from demonstrations incorporating expert knowledge \cite{Hester17}.

Further, we provide classic policy gradients (REINFORCE) \cite{Williams1992}, trust region policy optimization \cite{schulman2015trust}, and proximal policy optimization (PPO) \cite{schulman2017proximal} from the spectrum of policy-based methods, which all support categorical, continuous and bounded action spaces. It is worth pointing out that many new algorithms only modify classic Q-learning or policy gradients by slightly changing the loss functions, and implementing them only requires a few lines of code on top of existing TensorForce components.

\begin{lstlisting}[float,caption={Agent API example},belowskip=-6mm,label=li:tensorforce,moredelim={[is][emphstyle]{@@}{@@}}]
from tensorforce.agents import PPOAgent
# Create a Proximal Policy Optimization agent
agent = PPOAgent(
    states=dict(type='float', shape=(10,)),
    actions=dict(
        discrete_action=dict(type='int', num_actions=10),
        binary_action=dict(type='bool')
    ),
    network=[
        dict(type='dense', size=64),
        dict(type='dense', size=64)
    ],
    step_optimizer=dict(
        type='adam',
        learning_rate=1e-4
    ),
    execution=dict(type='single'),
    states_preprocessing= [dict(type='running_standardize')]
)
// Connect to a client
client = DBClient(host='localhost',port=8080)
while True:
    # Poll client for new state, get prediction, execute
    action = agent.act(state=client.get_state())
    reward = client.execute(action)
	
    # Observe feedback
    agent.observe(reward=reward, terminal=False)
\end{lstlisting}

\head{Example usage.} We illustrate how users might interact with the API in Listing \ref{li:tensorforce}.  Developers specify a configuration containing at least a network specification and a description of states and action formats. Here, a single state with 10 inputs and two separate actions per step, one boolean, one discrete with 10 options are required. Single-node execution is chosen, and incoming states are normalized via a state preprocessor. Crucially, a large number of commonly used heuristics is both optional and transparently configurable. 

Next, a PPO (a state-of-the-art policy optimization method, e.g. used in OpenAI's recent work on DOTA \cite{openaidota}) agent is created using the configuration, and a client is instantiated to interact with an example remote system which we desire to control. The agent can now be used  by retrieving new state signals from the client, which needs to map system state (e.g. load) to inputs, and requesting actions from the agent. The client must implement these actions by mapping numerical representations such as the index of a discrete action to a change in the system. Finally, the agent has to observe the reward to provide feedback to the agent. The agent will automatically trigger updates to the underlying TensorFlow graph based on algorithm semantics, e.g. episode based, batch-based, or time-step based. 

Developers are thus freed from dealing with low-level semantics of deep learning frameworks and can concentrate on mapping their system to inputs, rewards and actions. By changing a few lines in the configuration, algorithm, data collection, learning, or neural network logic can be fine-tuned. Finally, the JSON configurations can be conveniently passed to auto-tuners for hyper-parameter optimization.

\subsection{LIFT}
LIFT uses the declarative agent API and a small set of reusable components to realize three different execution modes which we describe in this section. 

\head{Pretraining.} In pretraining mode, LIFT does not interact with a system but is provided with a trace data source such as a comma separated file, a database table, or a distributed file system. LIFT parses and maps these to demonstrations (described in detail in section \ref{traces}), creates an RL agent supporting pretraining, and imports data. It then executes and monitors pretraining through evaluators, i.e. by validating model performance, and finally by serializing the model.

\head{Agent-driven.} In agent-driven or \textbf{active execution}, LIFT alternates between interacting with the system (i.e. the environment) and the RL agent via the TensorForce API. Here, execution time is almost exclusively governed by waiting on the environment, as we show in \S\ref{evaluation}. The RL libraries we surveyed typically only offer agent-driven execution (e.g. OpenAI baselines) where this execution is tightly coupled with reinforcement learning logic. This is because training common simulation tasks such as the Arcade Learning Environment \cite{Bellemare2013} can be effectively parallelized to hundreds of instances due to marginal computational requirements per simulator process. These highly parallel training procedures are economically impractical for users without data center scale resources, as learning to control data processing systems requires significant I/O and compute.

\head{Environment-driven.} In environment-driven execution or \textbf{passive execution}, LIFT acts as a passive service as control flow is driven by external workload, e.g. a benchmark suite executed against a database. For example, LIFT may open up a websocket or RPC connection to a monitoring service to receive real-time performance metrics. The LIFT controller then continuously maps incoming metrics to states, passes them to the agent, and executes the necessary configuration changes on the system. Passive execution is primarily intended for deployment of trained models which can optionally perform incremental updates. All execution modes share a common set of components which users need to implement for their given system to facilitate the parsing and serialization overhead necessary to interface a system.

First, a \textbf{schema} is used to programmatically construct the layouts of states, actions and rewards. For example, in our compound indexing case study, the input size to the neural network depends on the number of available query operators and unique fields in the database.  In our experience, successful application of RL initially requires frequent exploratory iterations over different state and action layouts. In LIFT, this is reflected by users implementing multiple exchangeable schemas. Downstream components for the execution modes use a schema to infer shape and type information.

Next, users implement a \textbf{model converter} as the central component for translating between RL model and controlled system via a small set of methods called throughout LIFT to i) map system output to agent states and agent actions (for pretraining), ii) map system output to rewards, and iii) map agent output to system configuration changes. LIFT's generic components for each execution mode then use converters to deserialize and parse log traces, and to perform offline (pretraining) and online (agent- or environment-driven) training. 

We summarize the idea behind LIFT as motivated by two observations. First, unlike common RL simulation tasks, controlling data processing systems requires separation of environment and RL agent due to different resource needs and communication patterns (e.g. access to system metrics through RPC or other protocols). Second, using RL in practical contexts currently requires a large amount of boiler-plate code as no standard tools are available. LIFT enables researchers to focus on understanding their state, action and reward semantics and express them in a schema and system model, which generate the respective TensorFlow graphs via the TensorForce API. In the following section, we explain the pretraining process on the indexing case study.

\head{Implementation.} We implemented our LIFT prototype in $\approx$10000 lines of Python code which includes components for our example case studies. In this work, no low-latency access is required (e.g. for learning to represent data structures as described by Kraska et al. \cite{Kraska2017}) but we may implement a C++ serving layer in future case studies. 

\section{Learning from traces}\label{traces}
\subsection{Problem setup}
We now illustrate the use of LIFT in an end-to-end example based on our compound database indexing application. In database management, effective query indexing strategies are crucial for meeting performance objectives. Index data structures can accelerate query execution times by multiple magnitudes by providing fast look-ups for specific query operators such as range comparisons (B-trees) or exist queries (Bloom filters). A single index can span multiple attributes, and query planners employ a wide range of heuristics to combine existing indices at runtime, e.g. by partial evaluation of a compound (multi-attribute) index. Determining optimal indices is complicated by space usage, maintenance cost, and the fact that indexing decisions cannot be made independently of runtime statistics, as index performance depends on attribute cardinality and workload distribution. In practice, indices are identified using various techniques ranging from offline tool-assisted analysis \cite{dbtuningsqlserver2005, Chaudhuri1998, Dageville2004} to online and adaptive indexing strategies \cite{Graefe2010, Idreos2011, Halim2012,Petraki2015}. Managed database-as-a-service (DBaaS) offerings sometimes offer a hybrid approach where indices for individual attributes are automatically created but users need to manually create compound indices.

We study MongoDB as a popular open source document database where data is organized as nested J/BSON documents. While a large body of work exists on adaptive indexing strategies for relational databases and columnar stores \cite{Petraki2015}, compound indexing in document databases has received less attention. Document databases are offered by all major cloud service providers, e.g. Microsoft's Azure CosmosDB offers native MongoDB support \cite{cosmosdb}, Amazon's AWS offers DynamoDB \cite{dynamodb}, and Google Cloud provides Cloud Datastore \cite{googlecloudstore}. The document database services we surveyed offer varying specialized query operators, index design, and query planners using different indexing heuristics. The aim of automatic online index selection is to omit this operational task from service users. We initially focus on common query operators available in most query dialects, as we plan to extend our work to other database layouts and  query languages. Table \ref{operator-table} gives an operator overview. In MongoDB, queries themselves are nested documents. 

\subsection{Modeling indexing decisions}
The MongoDB query planner uses a single index per query with the exception of $\$or$ expressions where each sub-expression can use a separate index. An index may span between $1$ and $k$ schema fields and is specified via an ordered sequence of tuples $(f_1, s_1),..,(f_n, s_n)$ where each tuple consists of a field name $f_i$ and a sort direction $s_i$ (ascending or descending). At runtime, the optimizer will use a number of heuristics to determine the best index to use. 

Via index intersection, the optimizer can also partially utilize existing indices to resolve queries. For example, \textit{prefix intersection} means that for any index sequence of length $k$, the optimizer can also use any ordered prefix of length $1..k-1$ to resolve queries which do not contain all $k$ attributes in the full index. Consequently, while the tuple ordering of the index does not typically matter for individual queries, the number of indices for the entire query set can be drastically reduced if index creation considers potential prefix intersections with other queries. Similarly, sort-ordering in indices can be used to sort query results via sort intersection in case of matching sort patterns. For example, an index of the shape $[(f_1, ASC), (f_2, DESC)]$ can be used to sort ascending/descending and descending/ascending (i.e. inverted) sort patterns, but not ascending/ascending or descending/descending. Based on these indexing rules, we define the following state, action, and reward model.
\begin{table}[t]
  \caption{MongoDB basic operator overview.}
  \label{operator-table}
  \centering
  \begin{tabular}{ll}
    \cmidrule{1-2}
   Operators     &  MongoDB operator \\
    \midrule
 $=$,$ >$, $\geq$, $<$, $\leq$, not in & $\$eq$, $\$gt$, $\$gte$, $\$lt$,$\$lte$, $\$nin$  \\
 and, or, nor, not & $\$and$ , $\$or$, $\$nor$, $\$not$     \\
  limit, sort, count  & $count()$, $limit(n)$, $sort(keys)$ \\
    \bottomrule
  \end{tabular}
\end{table}

\head{States.} Identifying the correct index for a query requires knowledge of the query shape, e.g. its operators and requested attributes. To leverage intersection, the state must also contain information on existing indices which could be used to evaluate a query. We parse queries via a tree-walk, strip concrete values from each sub-expression, and only retain a sequence of operators and attributes. If an index already exists on an attribute, we insert an additional token after the respective attribute to enable the agent to learn about index intersection and avoid adding unnecessary indices. For example, consider the simple following query counting entries with name "Jane":
\begin{alltt}
collection.find(\{\$eq: \{name: "Jane"\}\}).count() 
\end{alltt}
Assuming an ascending index on the \textit{name} field already exists, the tokenized query looks as follows (with EOS representing the end-of-sentence):
\begin{alltt}
[\$eq name IDX_ASC count EOS]
\end{alltt}
These tokens are then converted to integers using a word embedding as commonly used in natural language processing applications to map a discrete set of words to a continuous vector space \cite{Mikolov2013}. In practice, a maximum fixed input length is assumed and shorter inputs are padded with zeros.

\head{Actions.}
For every query we seek to output an index (or none)
spanning at most $k$ attributes where $k$ is a small number as indices covering more than 2-4 attributes are rare in practice. This is also because compound indices containing arrays, which require multi-key indices (each array element indexed separately), scale poorly and can slow down queries. Additionally, as discussed above, index intersection makes indices order- and sort-sensitive, thus requiring to also output a sort order per attribute in a multi-key index.

The action scheme should scale independently of the number of attributes in the document schema. Consider a combinatorial action model where the agent is modelled with one explicit action per attribute, and a separate action output per possible index-key. A 3-key index task on 10 attributes would already result in thousands of action options per step ($10^3*3=3000$) when including an extra action for the three possible sort patterns (both ascending/descending, descending-ascending, ascending-descending). This approach would not generalize to changing schemas or data sets. We propose a positional action model wherein the number of actions is linear in $k$. When receiving a query, we extract all query attributes and interpret an integer action as creating an index on the $ith$ input attribute, thus allowing the agent to learn the importance of key-order for prefix intersection. To distinguish sort patterns, we create an extra action per key (one ascending, one descending with ascending default). This results in $1+ 2k$ actions for a $k$-key index with one output for no-op. 

\begin{figure}[t] 
\centering
\includegraphics[scale=.7]{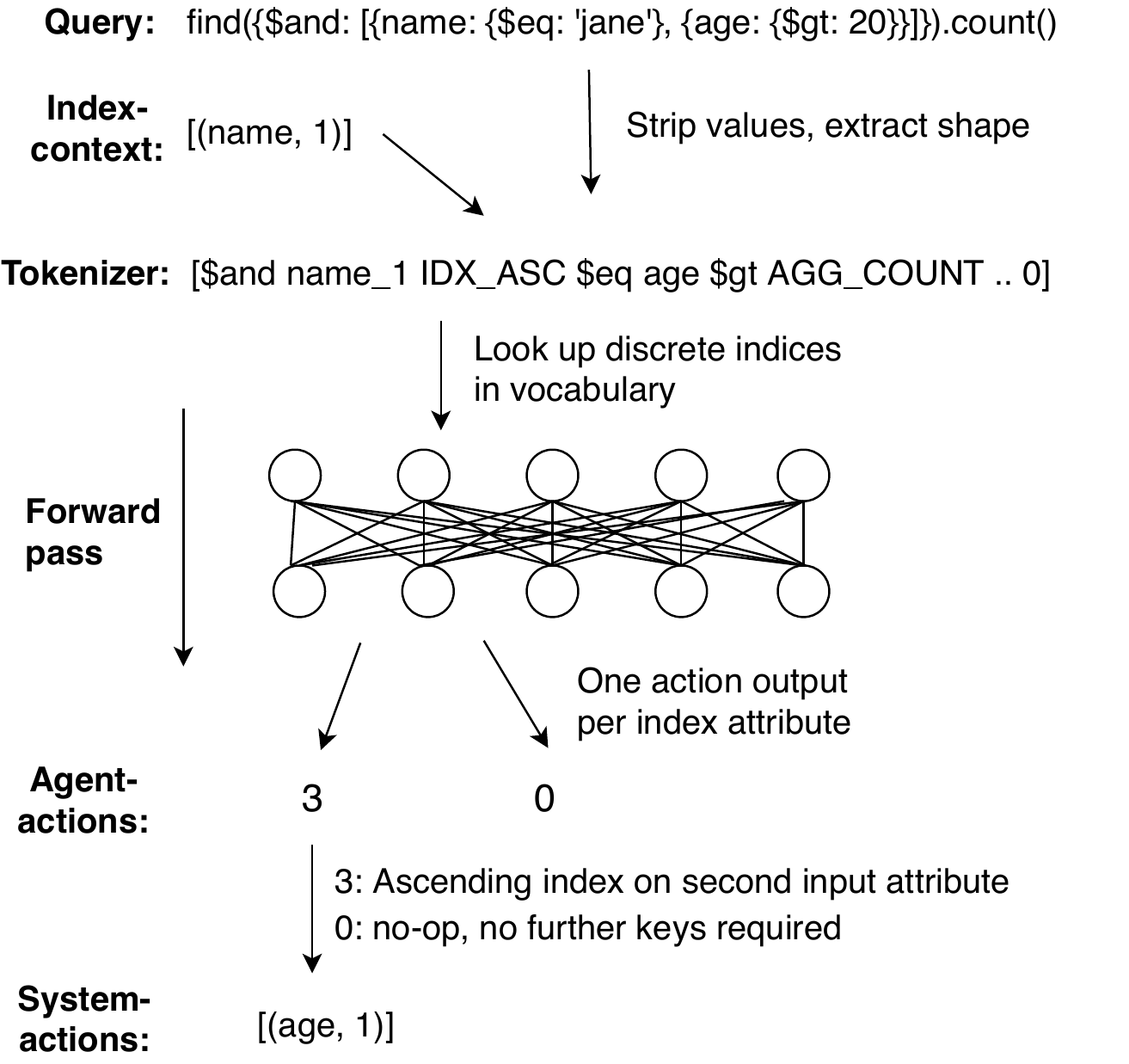}
\caption{State and action parsing scheme for the indexing case study.}
\label{fig:index-action-model}
\vspace{-5mm}
\end{figure} 
Figure \ref{fig:index-action-model} illustrates state and action parsing for $k=2$ and a simple query on \textit{name} and \textit{age} attributes. In the example, the \textit{name} field is already indexed so when the query is tokenized, a special index token ($IDX\_ASC$) is inserted to indicate the existing index. The tokenized sequence is mapped to integers via the embedding layer and passed through the network, which outputs $k$ integer actions. In the example, the agent decides to implement one additional single-key index by outputting 3 and 0, where 3 implies an ascending index on the second input attribute, and 0 is used for no-op if fewer than $k$ keys are required in the index.

\head{Rewards.}
The optimal indexing strategy is the minimal set of indices $\mathrm{I}$ meeting performance level objectives such as mean latency or 90th and 99th latency percentiles for a set of queries $\mathcal{Q}$. Let $t(q)$ be the time to execute a query $q \in \mathcal{Q}$ under an index set $\mathcal{I}$ and let $m(\mathcal{I})$ be the memory usage of the current index set. We set the reward $r(q)$ as the negative weighted combination of these to allow expressing trade-offs on memory usage against runtime requirements:

$r(q) = -\omega_1 m(\mathcal{I}) - \omega_2 t(q)$.

\subsection{Demonstrations and Pretraining}
We now describe the ideas behind learning from demonstrations as used in LIFT. Our approach is motivated by the observation that a human systems developer encountering a tuning problem can frequently use their expertise to come up with an initial heuristic. For example, in the indexing problem, a database expert can typically determine an effective configuration for a given application within a reasonable time frame (e.g. a few hours) with access to profiling tools. Distilling this intuitive expertise into a fully automated approach is difficult, and simple heuristics may perform well in small scenarios but fail at scale. Moreover, as discussed in \S \ref{practical}, training a RL model from scratch is expensive and difficult, while refining a model pretrained from not necessarily fully correct demonstrations may be more effective. We hence argue for an approach that leverages pre-existing domain knowledge by initializing training from demonstrations.

\head{Demonstration data.} In the indexing task, demonstrations may exist in the form of:
\begin{enumerate}
\item Query logs from applications configured by a database administrator where indices are assumed to be correct, where correctness implies fully meeting service level objectives (not necessarily being optimal).
\item Query logs from applications where indices were created using any heuristic understood to be sub-optimal and not necessarily meeting service objectives.
\item Queries and index pairs for which no runtime data is available, e.g. procedurally generated examples with either manually or heuristically chosen index recommendations (both correct and imperfect).
\end{enumerate}
The key difference between (1) and (2) is that when encountering a query for which an imperfect demonstration was available during pre-training, we do not mind testing other choices while this is unnecessary if a demonstration was optimal for the query given. This confidence must be reflected in the pretraining procedure. Further, the difference between (1), (2) and (3) is that in the latter, no reward is available without creating indices and measuring queries. Note that the key difference between demonstrations and simulation in our applications is the absence of information on system dynamics (i.e. state transitions). 

A simulator for query indexing would provide insights into how addition and removal of an index affects performance. In contrast, a demonstration extracted from the slow query log of a database indicates how fast a query performed using the index chosen by the query planner, but not how much faster the index was versus not using an index, or a different index. We make use of all demonstration types but focus on (2) and (3), as we could not obtain existing traces from expert-configured systems and thus had to manually tune configurations.

\head{Algorithm.}
Hester et al. have described an algorithm to perform Deep Q-learning from such expert demonstrations (DQfD) using the example of Atari games \cite{Hester17}. In their work, an agent is trained until sufficient performance, and then games played by that agent are given as demonstrations to a new agent. DQFD works by assigning an 'expert margin' to demonstration actions by extending double Q-learning  \cite{double_dqn}, a Q-learning variant which corrects biased Q- estimates in the original DQN by decoupling action selection and action evaluation. 
Specifically, the double DQN loss
\begin{align}
 J_{DQ}(Q)= (R(s,a) +\gamma Q(s_{t+1},a^{max}_{t+1};\theta')- Q(s,a;\theta))^2
\end{align}
where 
\begin{align}
a^{max}_{t+1} = argmax_a Q(s_{t+1},a;\theta)
\end{align}
uses the target network (as explained in \S \ref{background}, parametrized by $\theta'$)) to evaluate the action selected using the training network (with parameters $\theta$). This is combined with another expert loss function $J_E$:
\begin{align}
J_E(Q) =\max_{a \in A}[Q(s, a) + l(s, a_E, a)] - Q(s,a_E)
\end{align}
Here, $l(s,a_E,a)$ is a function which outputs 0 for the expert action, and a margin value $>0$ otherwise. We convey the intuition of this loss function by recalling the action selection mechanism in Q-learning. 

Recall that $Q(s,a,\theta)=\mathbb{E}[R_t|s_t=s,a]$, i.e. the expected returns from taking a decision $a$ in state $s$. At each step, the neural network (parameterized by $\theta$) used to approximate $Q$ outputs Q-values for all available actions and selects the action with the highest Q-value. By adding the expert margin to the loss of Q-values of \textit{incorrect actions}, the agent is biased towards the expert actions as a difference between expert actions and other actions of at least the margin is enforced \cite{Piot2014}. The DQFD-agent keeps a separate memory of these expert demonstrations which are first used to pretrain the agent, then combined with new online experiences at runtime so that the agent keeps being 'reminded' of demonstrations.

What does the choice of $l(s,a_E,a)$ imply for imperfect or noisy demonstrations? A large margin makes it difficult to learn about any better actions in a given state because even if, via exploration, a different action is selected and yields a higher return, an update may not change Q-values of better action beyond the margin. Second, the DQfD loss only enforces a difference in Q-values between demonstrated action and all other actions; no assumptions are made about the relationship between non-expert actions (e.g. second highest, third highest Q-value). This behavior is desirable in the indexing example because even semantically similar indices (e.g. different order, partially covering same fields) can result in much worse performance than the demonstrated index, so we initially do not want to express any preference on non-demonstrated indices. Consequently, we choose a very small margin $\leq 0.1$ which in practice results in a pre-generated model which initially only slightly favors the demonstrated action.

\begin{algorithm}
\begin{algorithmic}

\State Initialize $agent$ with demo-model and demo-data $D$
\State Initialize LIFT $system\_model$, $model\_converter$ 
\State Load application queries $\mathcal{Q}_{test}$
\State // Fixed time budget or until objectives met
\For{ $i=1, N$}
\State $\mathcal{I}_{test} \gets \emptyset$, clear index set in DB
\For{ $q$ in $\mathcal{Q}_{test}$}
\State // Tokenize, include existing indices 
\State $s(q) \gets model\_converter.to\_agent\_state(q, \mathcal{I}_{test})$
\State $index \gets agent.act(s(q))$
\State // Create index, execute query
\State $m(\mathcal{I}_{test}) \gets system\_model.act(index)$
\State $t(q) \gets system\_model.execute(q)$
\State // Compute reward from runtime and size
\State $r_q \gets -\omega_1 m(\mathcal{I}_{test}) - \omega_2 t(q)$
\State $agent.observe(r_q)$
\State Add $index$ to $\mathcal{I}_{test}$
\EndFor
\EndFor
\State // Final evaluation, create best $\mathcal{I}_{test}$:
\State // Measure final size $m(\mathcal{I}_{test})$, run $\mathcal{Q}_{test}$ 
\end{algorithmic}
\caption{Online training procedure.}
\label{training-procedure}
\end{algorithm}

\subsection{Putting it all together.}
Algorithm \ref{training-procedure} shows pseudo-code for the online training procedure. Following pre-training on the demonstration data set, we start LIFT in online mode, initialize an agent with the demo model, and load the demo data. We then begin the episodic training procedure on a new set of queries $\mathcal{Q}_{test}$ we want to index. In each training episode, all indices are first removed from the database. Then, each query $q$ (sorted by length to improve intersection) is tokenized and the suggested index created. Recall that the tokenization includes the current index set $\mathcal{I}$ for the agent to learn the impact of existing indices. The size of the index set $m(\mathcal{I})$ and the runtime of the query $t(q)$ are used to inform the reward of the agent. For direct search tasks like indexing, we keep the list of index tuples associated with the highest reward during training. In the final evaluation, we recreate these indices and then run all queries 5 times on the full index set. For dynamic tasks where the agent is invoked repeatedly at runtime, we simply export the trained model which can then be used to control a system.

\section{Evaluation}\label{evaluation}
\subsection{Aims}
We evaluate our LIFT prototype through two case studies: 1) the indexing case study in which we minimize latency and memory usage by learning compound index combinations, and 2) the stream processing resource management case study in which we tune latency by setting parallelism levels under a varying workload. In both case studies, we used LIFT to implement a controller, manage demonstration data, and interact with the system. The difference is that the indexing task is an offline optimization (index set is determined once, then deployed), while in the stream processing task we use a controller at runtime to react to varying workloads. The evaluation focuses on evaluating the utility of LIFT and TensorForce to solve data management tasks, and on understanding the impact of learning from demonstrations to overcome long training times.
\newcommand{\customtextwidth}{.24}
\begin{figure*}[ht]
\centering
\begin{subfigure}[t]{.24\textwidth}
\includegraphics[ clip, scale=\customtextwidth]{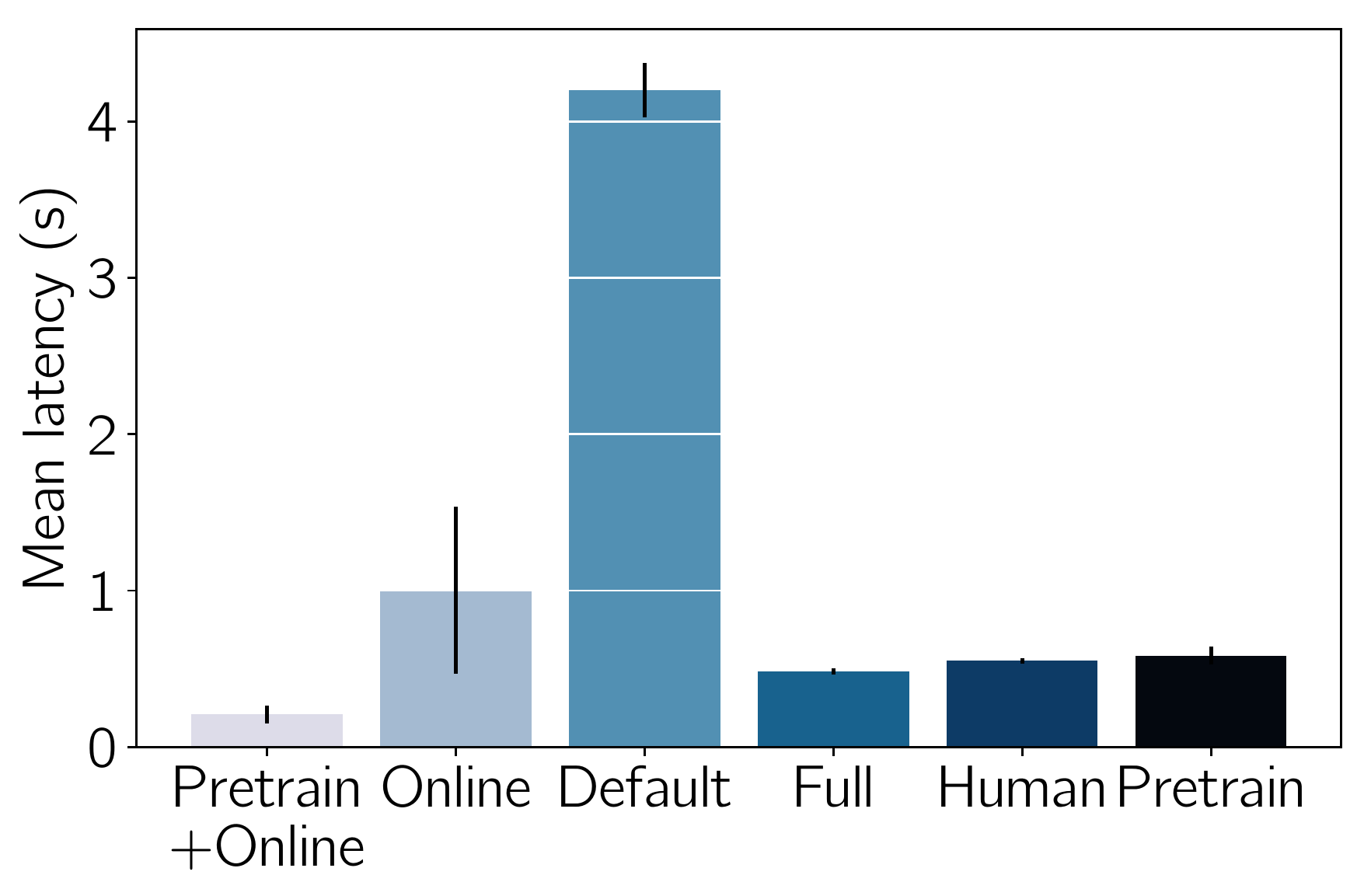}
\caption{\label{fig:latency-mean} Mean latency.}
\end{subfigure}
\begin{subfigure}[t]{.24\textwidth}
\includegraphics[ clip, scale=\customtextwidth]{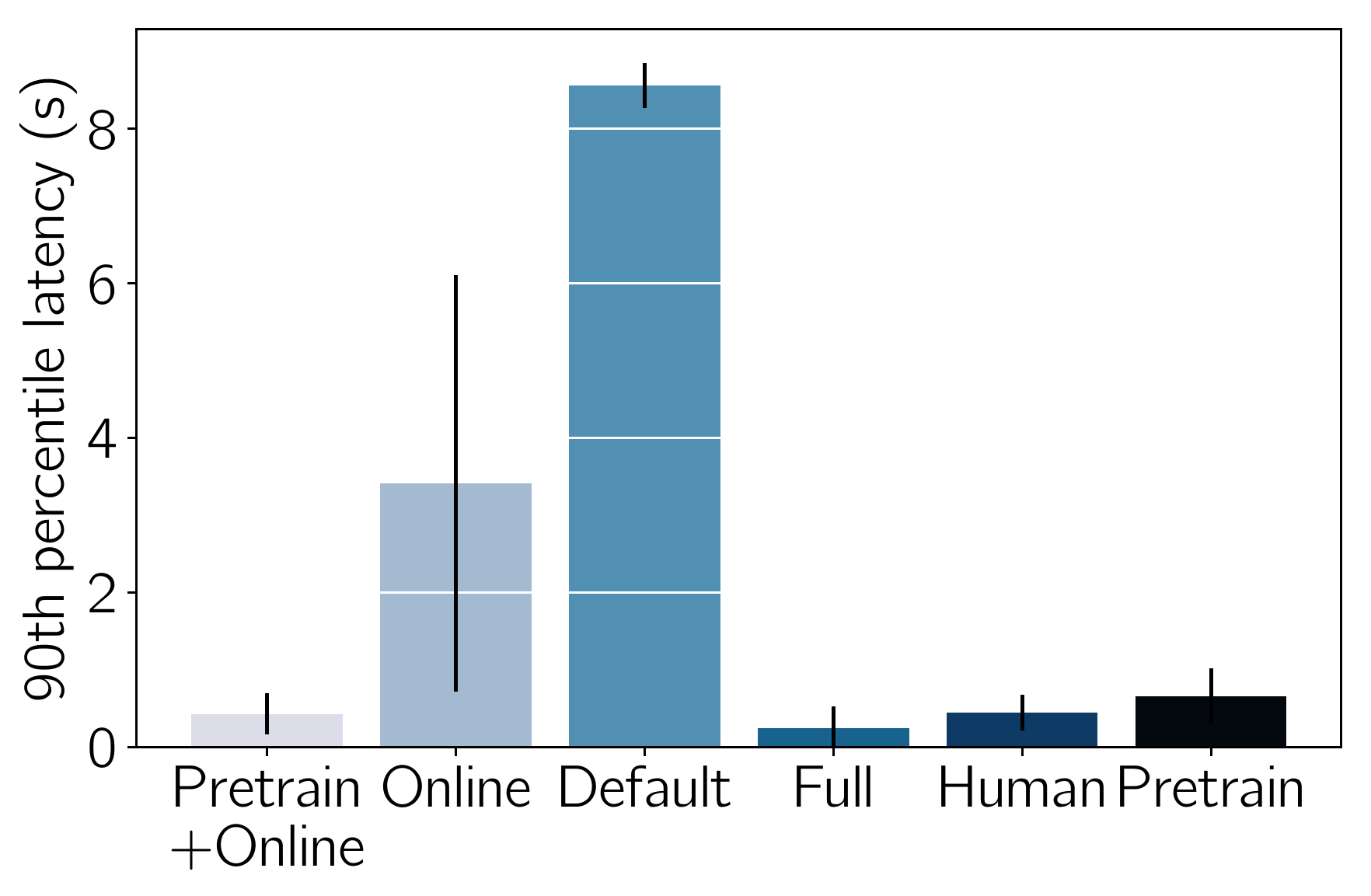}
\caption{\label{fig:latency-90} 90th pct. latency.}
\end{subfigure} 
\begin{subfigure}[t]{.24\textwidth}
\includegraphics[ clip, scale=\customtextwidth]{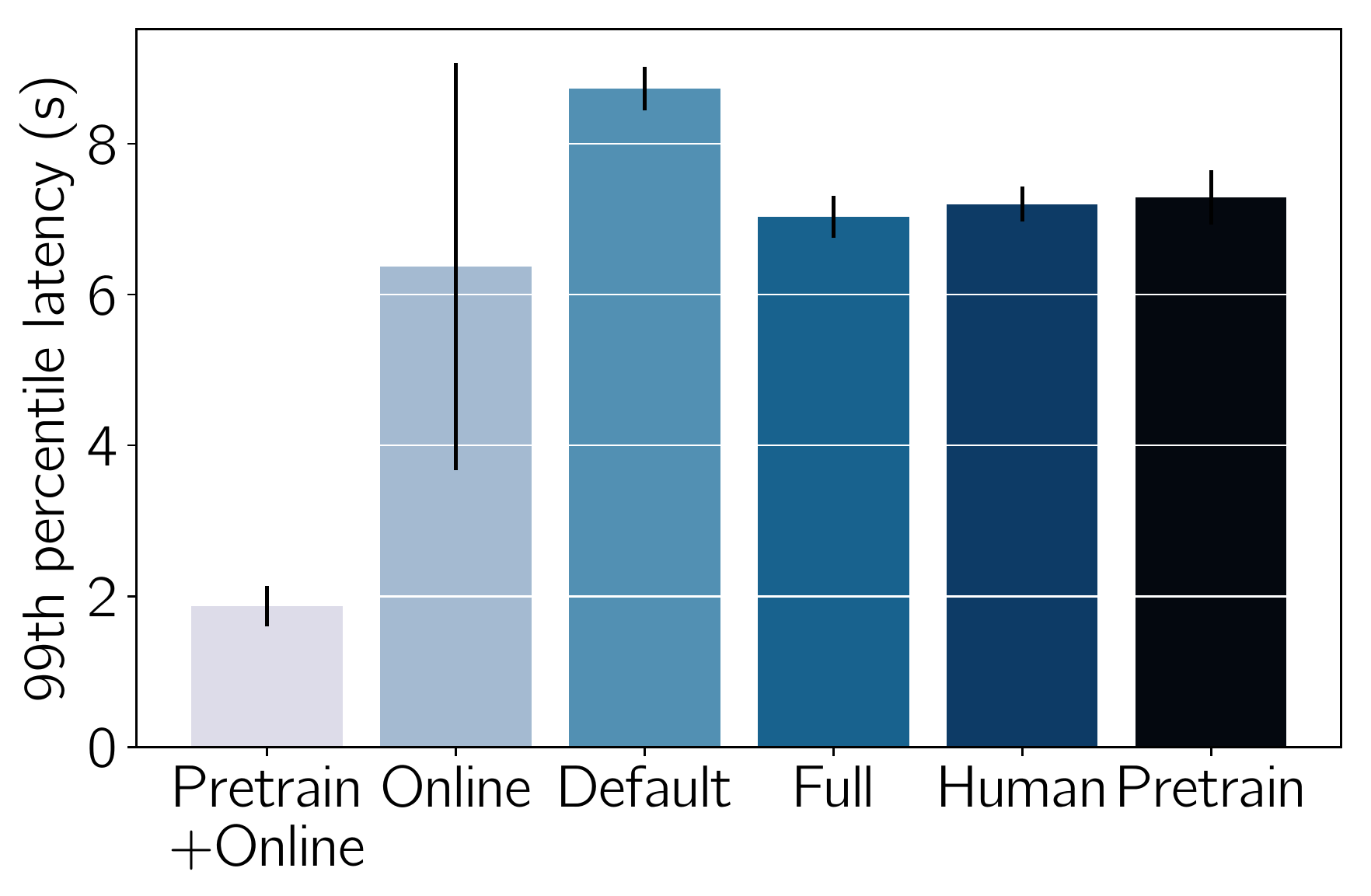}
\caption{\label{fig:latency-99}99th pct. latency.}
\end{subfigure} 
\begin{subfigure}[t]{.24\textwidth}
\includegraphics[ clip, scale=\customtextwidth]{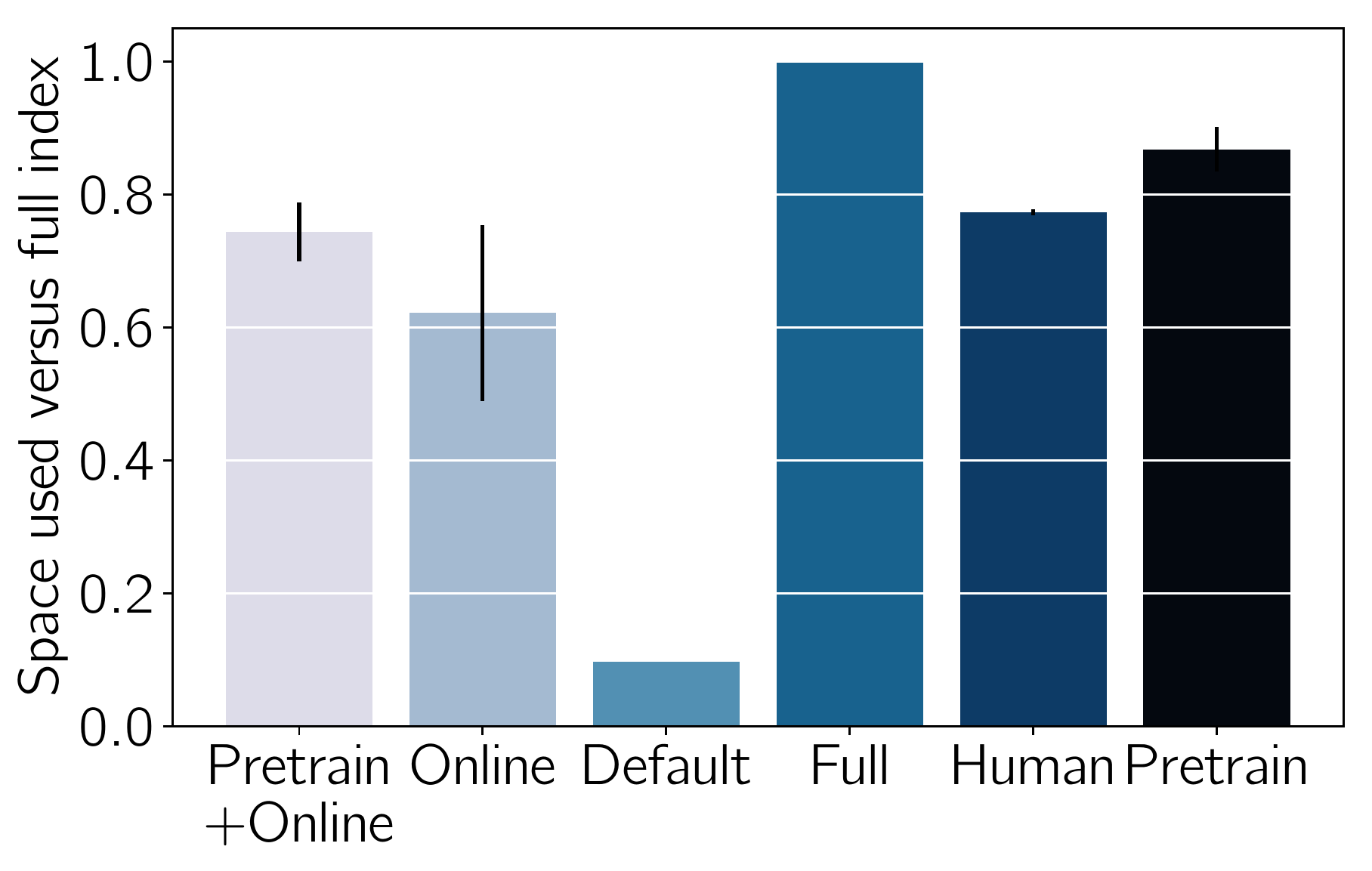}
\caption{\label{fig:index-size} Normalized index size.}
\end{subfigure}
\\
\begin{subfigure}[t]{.24\textwidth}
\includegraphics[clip,scale=\customtextwidth]{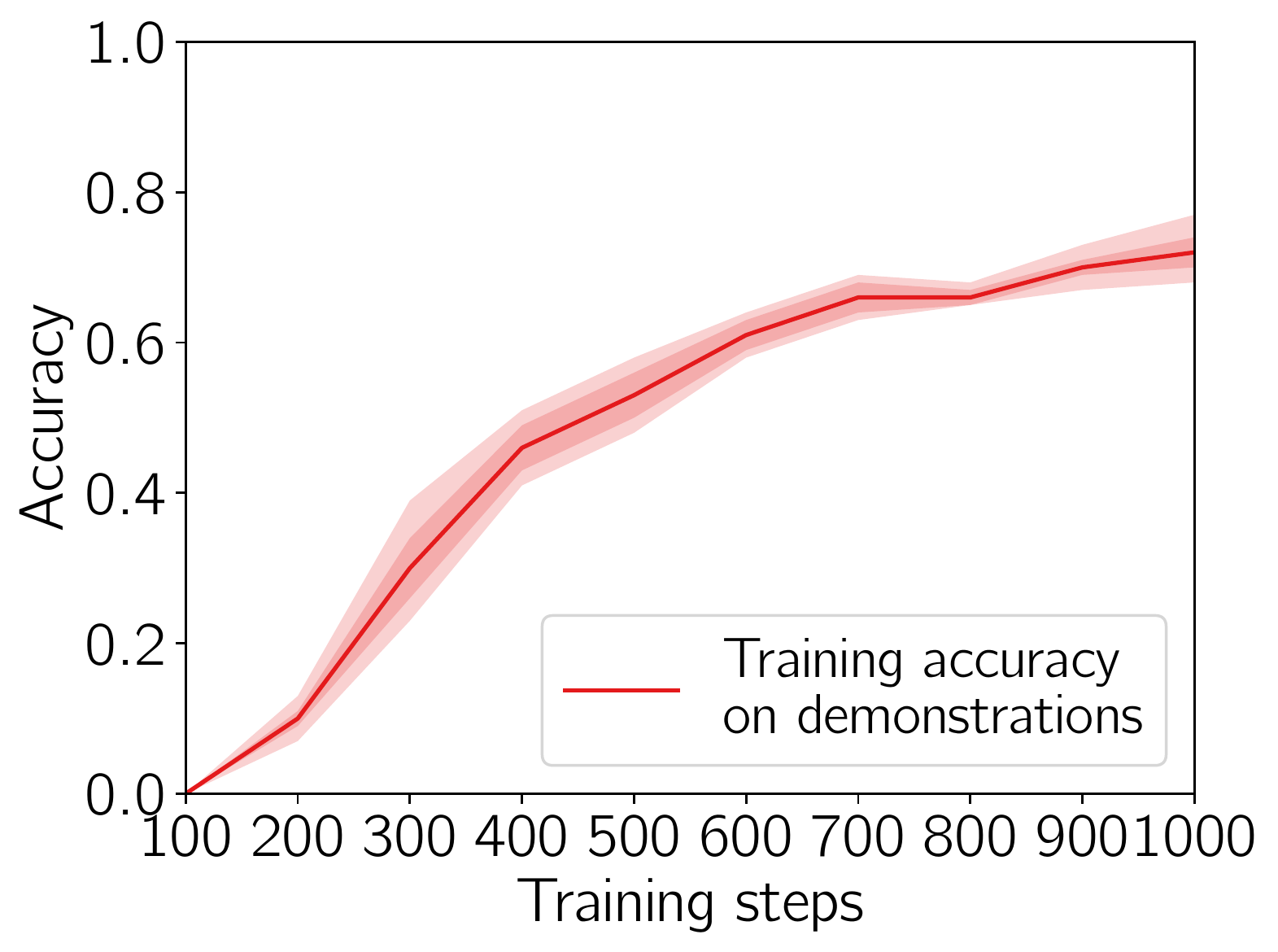}
\caption{\label{fig:pretraining-curve} Pretraining accuracy.}
\end{subfigure}
\begin{subfigure}[t]{.24\textwidth}
\includegraphics[ clip, scale=\customtextwidth]{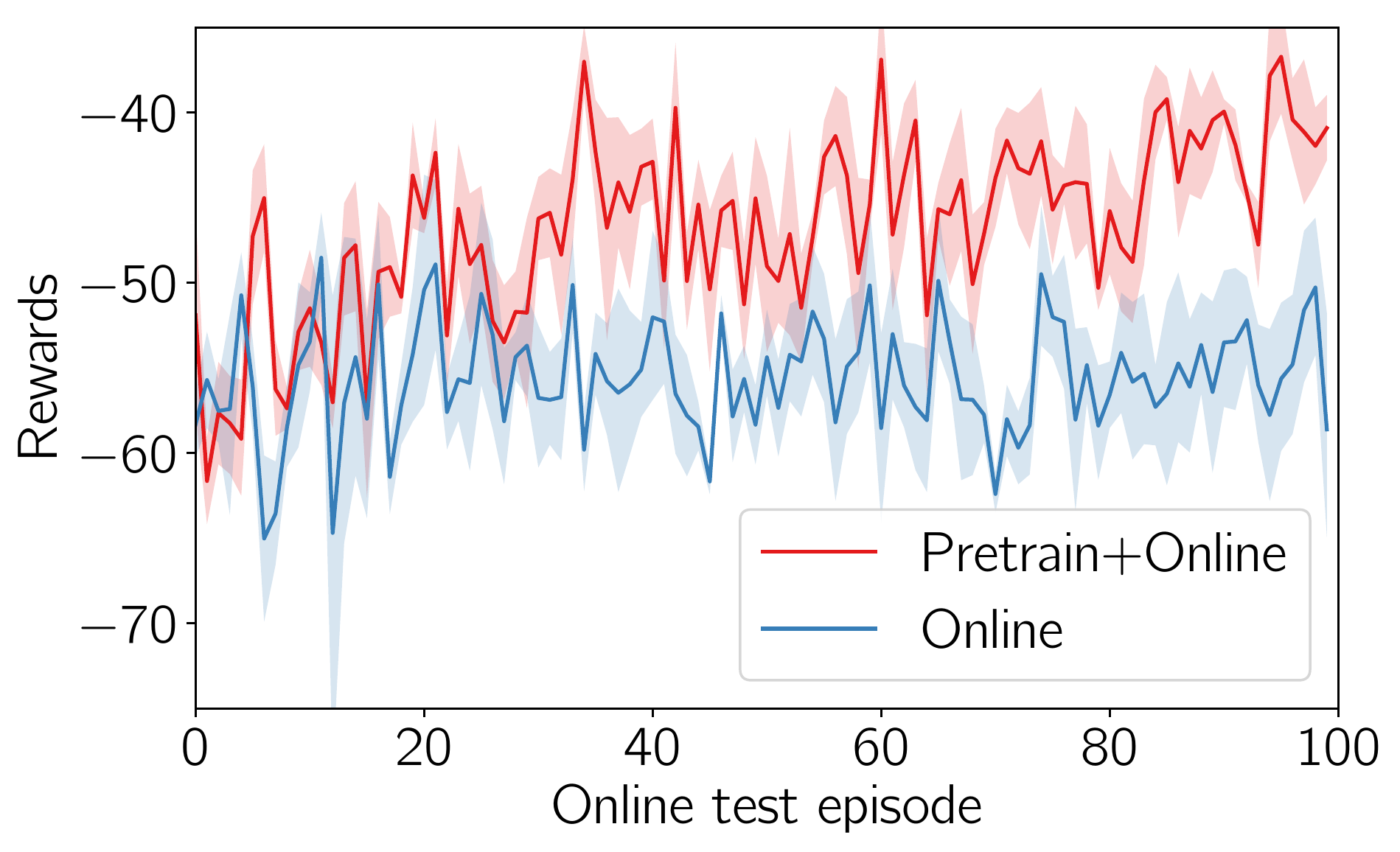}
\caption{\label{fig:online-curve} Online adaption rewards.}
\end{subfigure} 
\begin{subfigure}[t]{.24\textwidth}
\includegraphics[ clip, scale=\customtextwidth]{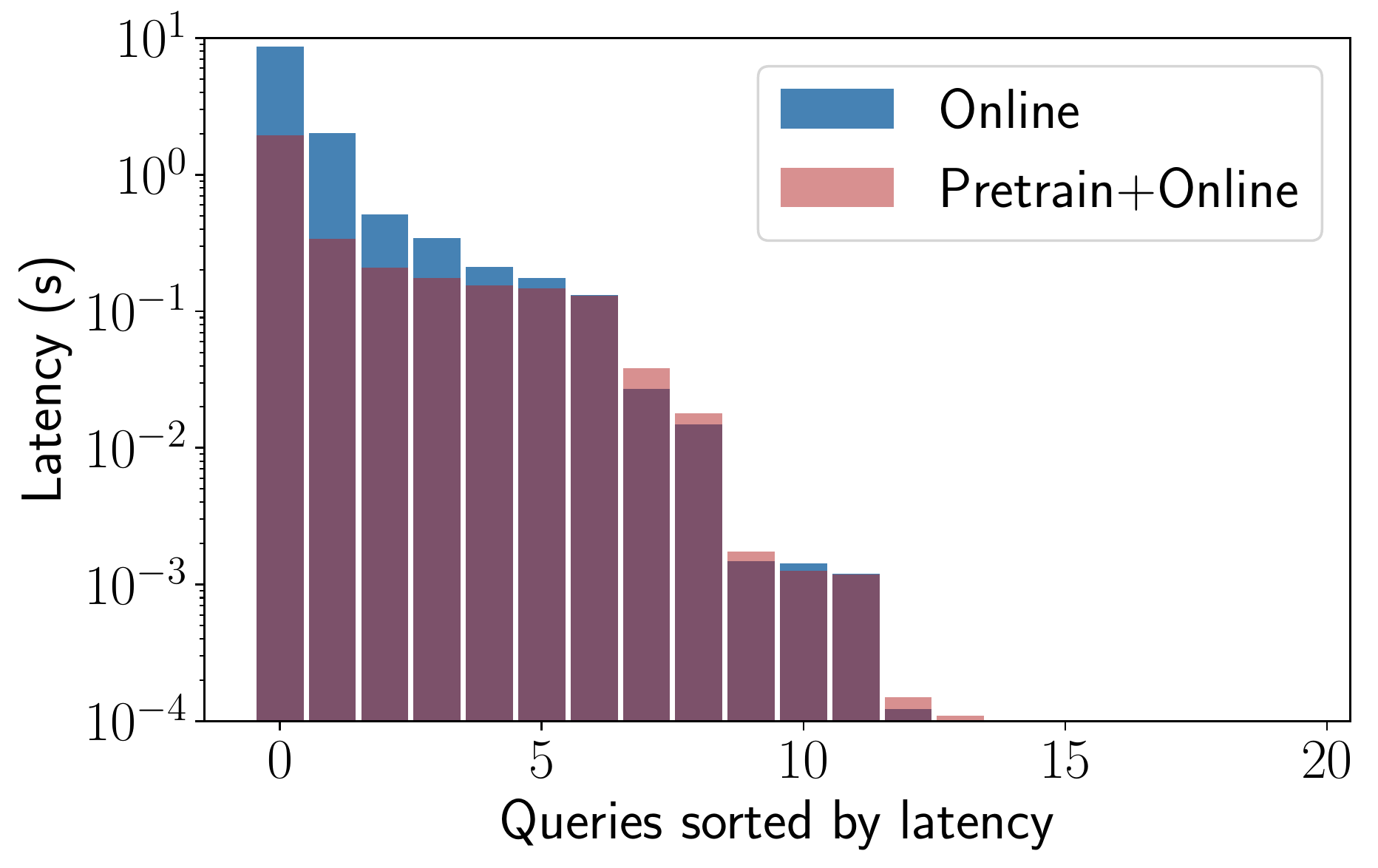}
\caption{\label{fig:latency-overlap}Per-query latency.}
\end{subfigure} 
\begin{subfigure}[t]{.24\textwidth}
\includegraphics[ clip, scale=\customtextwidth]{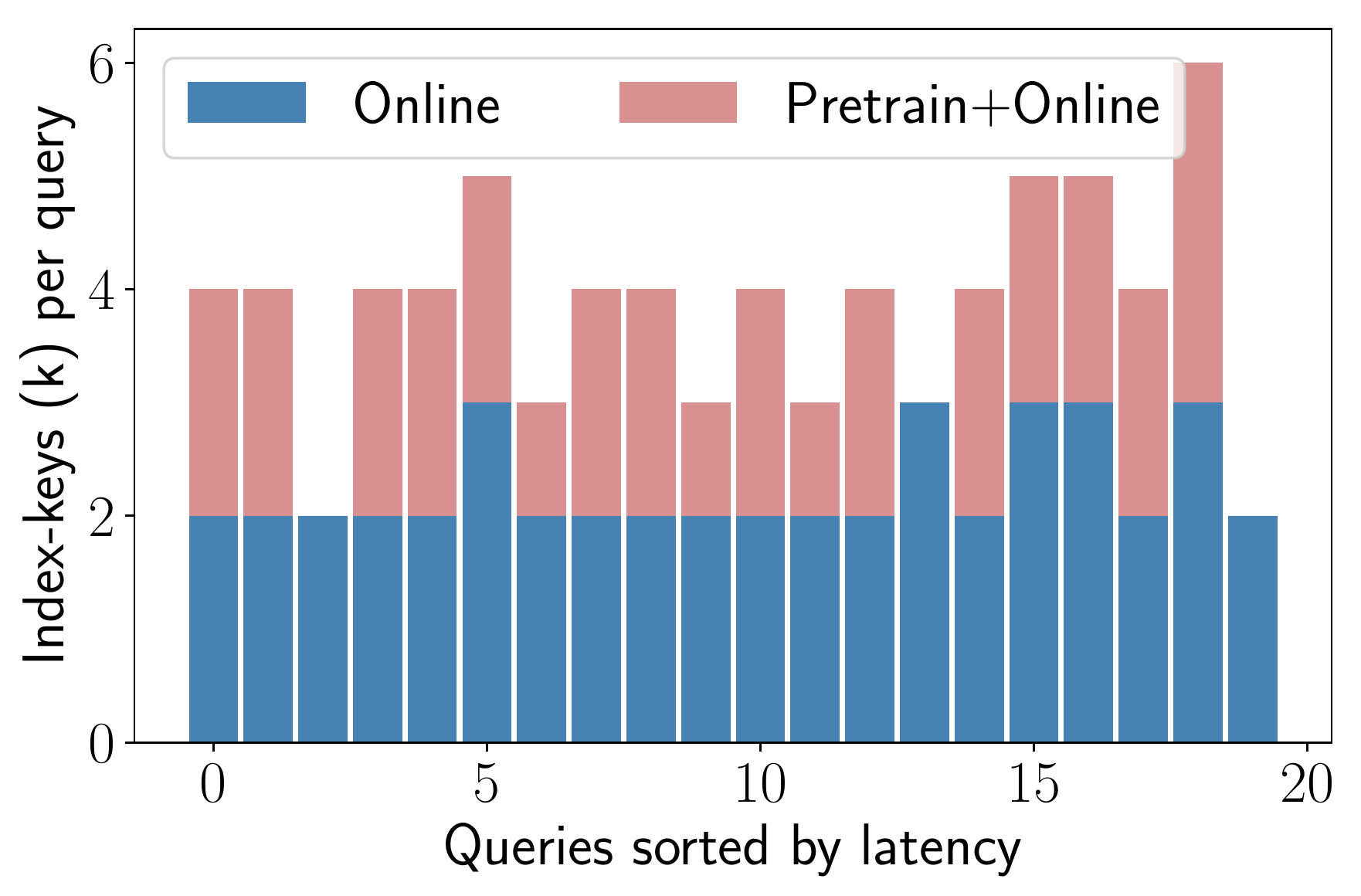}
\caption{\label{fig:index-breakdown}Per query index keys.}
\end{subfigure} 
\caption{Performance evaluation on the IMDB data set.}
\end{figure*}
\subsection{Compound indexing}
\head{Setup.}
We evaluate the indexing task both on a real-world dataset (IMDB \cite{imdb}) and using synthetic queries and data. The synthetic query client is based on the YCSB benchmark \cite{CooperSilbersteinTamEtAl2010}. YCSB generates keys and synthetic values to evaluate cloud database performance via a set of common workload mixtures but has no provisions for complex queries. We implemented a YCSB-style client and workload generator targeting secondary indexing. The client is configured with a schema containing attribute names and types. The workload generator receives a query configuration containing valid query operators, maximum allowed operator degrees, query height, and distribution of aggregation operators. It can then generate a specified number of queries index suggestions based on provided rules. All experiments were run on a variety of commodity server class machines (e.g. 24 cores (Xeon E5-2600) and 198 GB RAM) and using MongoDB v3.6.4. 

Queries and demonstrations are imported into LIFT's pretrain-controller which instantiates a DQfD agent and parses queries and demonstrations to states and actions as described before. We then run a small number of pretraining steps until the agent has approximately adopted the rule. We use batch sizes of 32 queries on a neural network with 1 embedding layer and 1 dense layer with 128 neurons each. Learning is executed using an adaptive moment optimizer (Adam \cite{KingmaBa2014}) with a learning rate of $0.0005$, and an expert margin of $0.1$. To refine the pretrained model, we restart LIFT in online mode and train as described in Algorithm 1. The max number of attributes per index was $k=3$. 

\head{Indexing baselines.}
We consider human and rule-based baselines due to a lack of standard tools for automatic indexing in document stores. The first rule-based strategy we use to generate demonstrations is full indexing (\textit{Full} in the following) wherein we simply create a compound index covering all fields in a query (respecting its sort order), thus ensuring an index exists for every query. In the synthetic regime, where query shapes are re-sampled every experiment to evaluate generalization to different query sets, human baselines were uneconomical, and we experimented with other rule-based baselines. Partial indexing (\textit{Partial} hereafter) attempts to avoid unnecessary indices by considering existing indices on any attribute in a query, and only covering unindexed fields. Note that we do not claim these to be the most effective heuristics but merely initial guidance for a model. We also experimented with a rule based on operator and schema hints but this frequently did not perform well due to unforeseen edge cases. We refer to the following modes in the evaluation: no indexing (\textit{Default}), online learning from scratch without pretraining (\textit{Online}), pretraining without online refinement (\textit{Pretrain}, online learning following pretraining (\textit{Pretrain+Online}), human expert (\textit{Human}and the two baselines described above. 

\head{Basic behaviour.}
We first show results on the publicly available internet movie database (IMDB) datasets \cite{imdb}. We imported datasets for titles and ratings (\textit{title.akas}, \textit{title.basics}, \textit{title.ratings}) comprising $\approx$10 million documents. We manually defined a representative set of 20 queries such as 'How many comedies with length of 90 minutes or less were made  before 2000?'. For this fixed set, we compared our method to human expert intuition. Using human baselines (which are common in deep learning tasks) in data management is difficult due to inherent bias and prior knowledge on experiment design. Generally, a human expert can identify effective indices for a small query set given unlimited trials to refine guesses. For a more interesting comparison, we hence devised a single-pass experiment where the expert was allowed to observe runtimes on the full indexing baseline and subsequently tried to estimate an index per query.
\begin{figure*}[ht]
\centering
\begin{subfigure}[t]{.24\textwidth}
\includegraphics[ clip, scale=\customtextwidth]{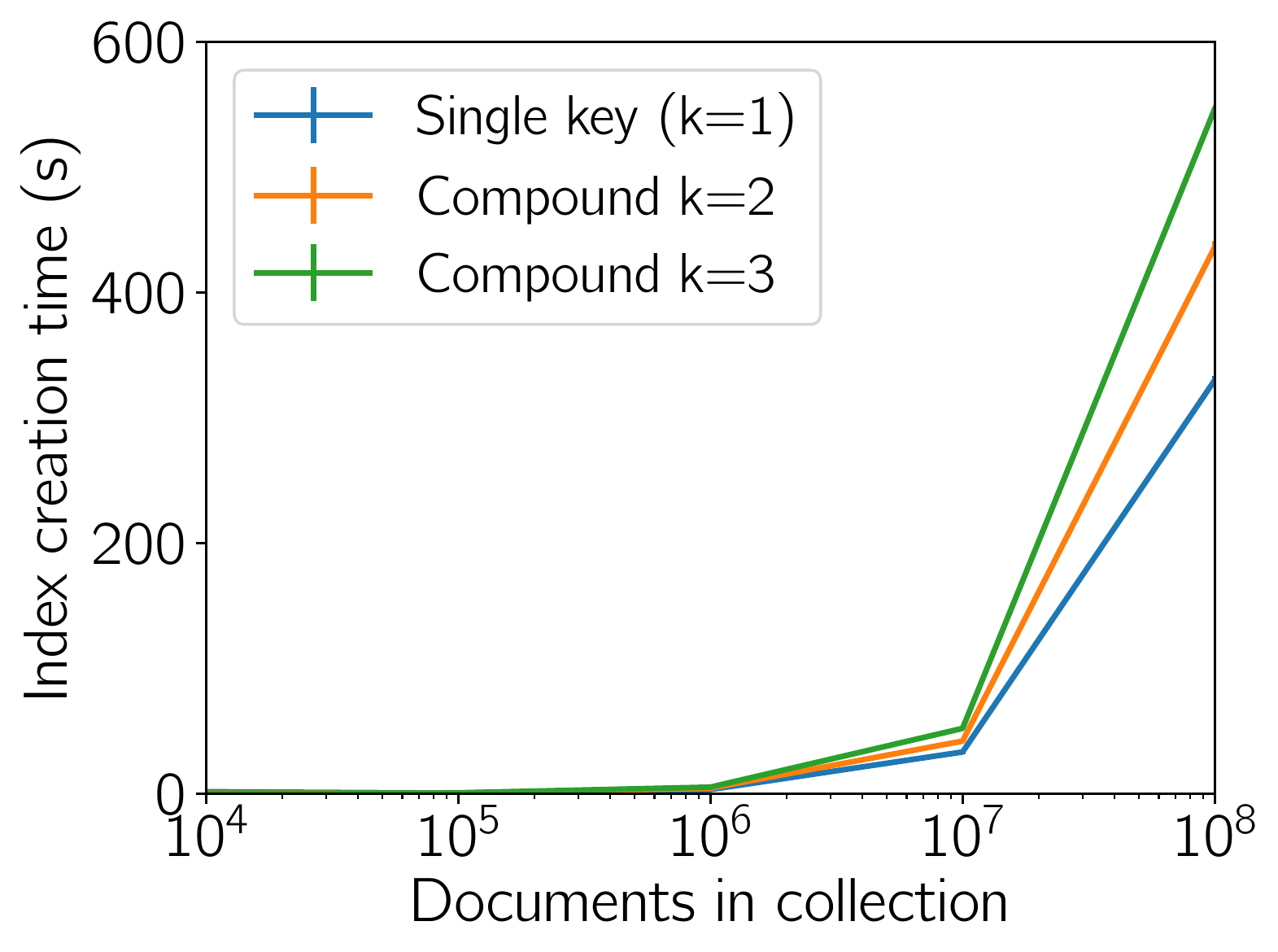}
\caption{\label{fig:runtimes} Index creation times.}
\end{subfigure} 
\begin{subfigure}[t]{.24\textwidth}
\includegraphics[ clip, scale=\customtextwidth]{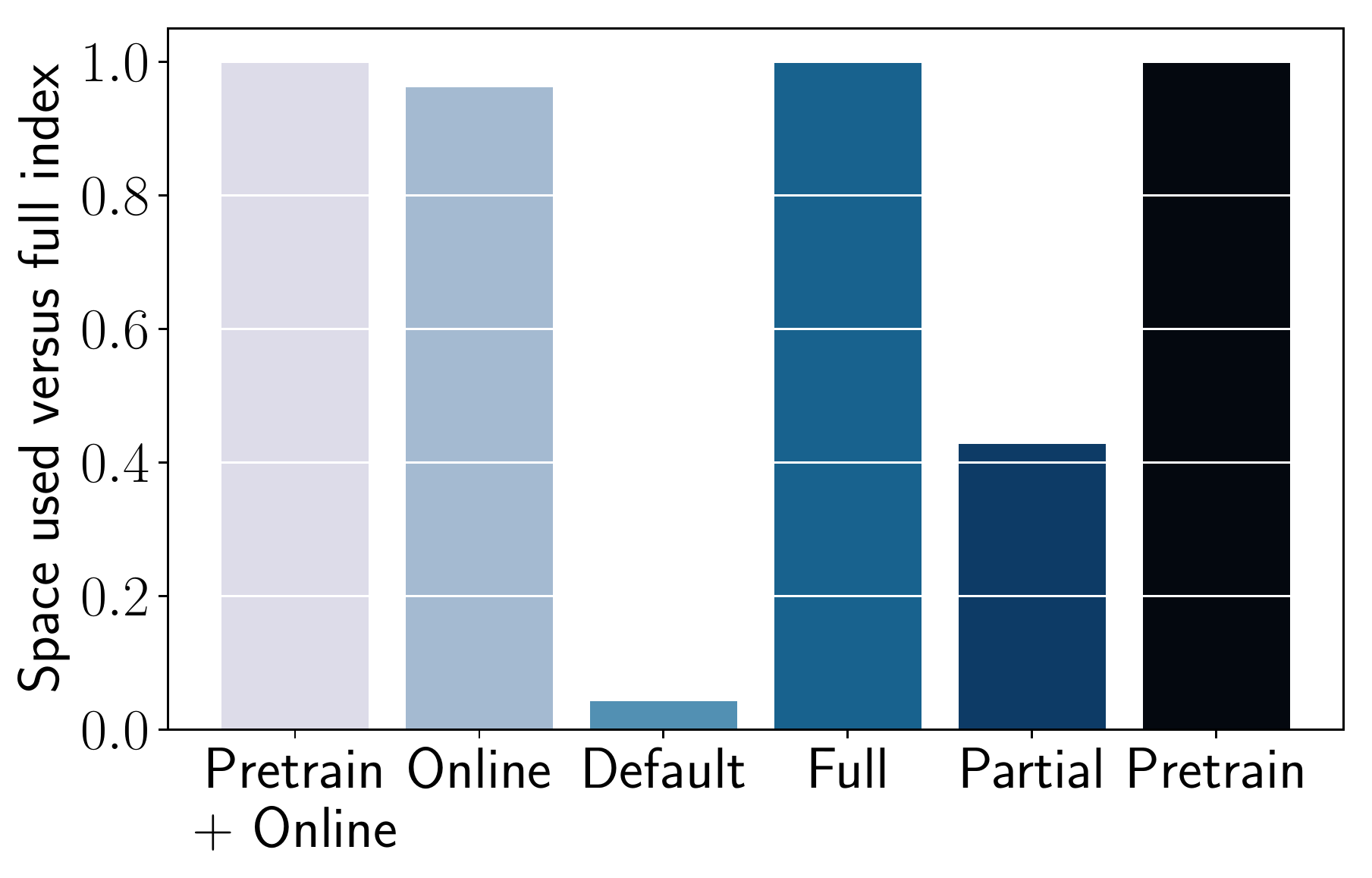}
\caption{\label{fig:scalability-size} Index size in scale-up.}
\end{subfigure}
\begin{subfigure}[t]{.24\textwidth}
\includegraphics[ clip, scale=\customtextwidth]{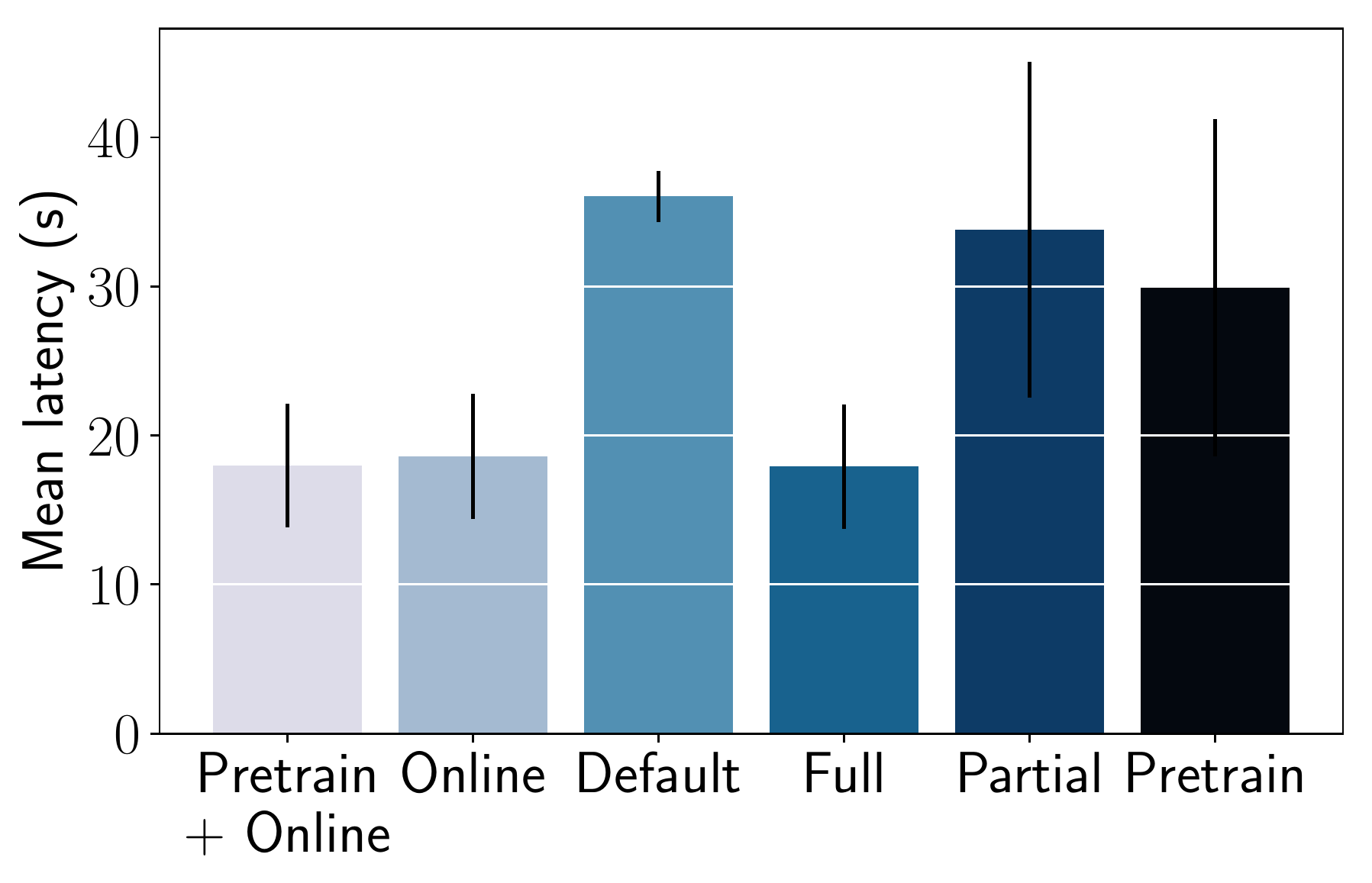}
\caption{\label{fig:scalability-mean-latency} Test set mean latencies.}
\end{subfigure} 
\begin{subfigure}[t]{.24\textwidth}
\includegraphics[ clip, scale=\customtextwidth]{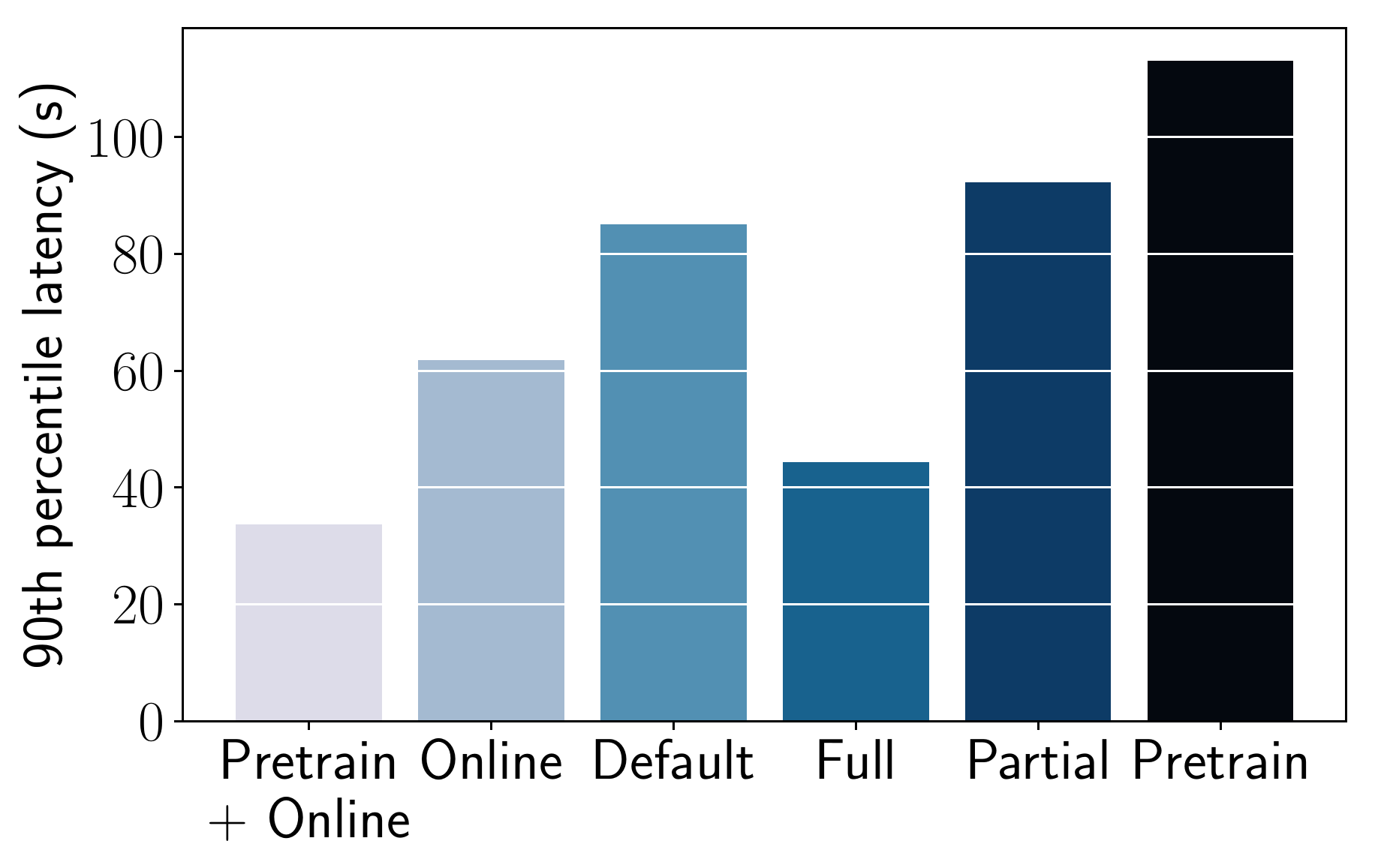}
\caption{\label{fig:scalability-90-latency} Test set 90th pctl. latencies.}
\end{subfigure} 
\caption{Scalability generalization analysis using the synthetic query client. Learning was performed on 10 million documents, a new set of test queries was evaluated on 100 million documents.}
\end{figure*}
Figures \ref{fig:latency-mean}, \ref{fig:latency-90} and \ref{fig:latency-99} give mean, 90th and 99th latencies respectively on the final evaluation in which each query is executed 5 times (final results averaged over five trainings with different random seeds). The combined \textit{Pretrain+Online} strategy outperforms other methods significantly, in particular improving mean latency by $57\%$ and $62\%$ against \textit{Full} and \textit{Human} respectively and by $74\%$ on 99th percentile latency against both. Differences in 90th percentile latencies were within standard error. In the human experiment, the expert attempted to reduce indices by leveraging intersection, creating only 14 indices versus 15 on average for \textit{Pretrain+Online}. The size of the created indices was however (marginally) bigger compared to \textit{Pretrain+Online} (as shown in Figure \ref{fig:index-size}) which achieved on average $25\%$ index size improvement against the \textit{Full} strategy. Note that MongoDB always creates a default index on the \textit{\_id} attribute so the default size is not zero. We normalize sizes against the size of the full index to evaluate improvement. The expert's attempt to exploit intersection also underestimated the necessity of compound indices for some queries. The outcome illustrates the difficulty of solving the task without iterative manual analysis.

\textit{Pretrain} and \textit{Online} can perform similar to full indexing. The performance of \textit{Pretrain} (in its degree of similarity to \textit{Full}) depends on whether pretraining is continued until the rule is fully adopted. We found early stopping at $70-80\%$ accuracy to be effective when using our imperfect rules to avoid overfitting (Figure \ref{fig:pretraining-curve}. \textit{Online} can sometimes find good configurations but tends to perform significantly worse than \textit{Pretrain+Online} in mean reward due to random initialization, as seen in Figure \ref{fig:online-curve} which shows reward curves (i.e. combined size and latency) and 1 $\sigma$ confidence intervals over 5 runs. We breakdown individual queries and indices for \textit{Pretrain+Online} and \textit{Online}, i.e. standard RL. In Figure \ref{fig:latency-overlap}, we sort queries of one experiment by latency and show runtime differences (n.b. log scale), and in figure \ref{fig:index-breakdown} the number of keys in the index decision (0-3) for the query (stacked on top of each other). Performance differences are concentrated in the five slowest queries with the rest being effectively indexed by both strategies. Comparing keys used per query also shows that \textit{Pretrain+Online} created indices systematically spanning fewer keys than \textit{Online}. This does not necessarily imply smaller total index size depending on the attributes indexed but indicates more effective intersection.

\begin{table}[t]
  \centering
  \begin{tabular}{lll}
    \cmidrule{1-3}
      Workload means  & Total time   &  Pct. \\
    \midrule
    Waiting on system & 64869 s ($\pm$ 4403 s) & 97.8 \% \\
    Agent interaction/evaluation & 1446 s ($\pm$ 218 s) & 2.2 \%     \\
    Mean episode duration & 663 s ($\pm$ 42 s) &  n/a    \\
    Min episode duration & 419 s ($\pm$ 88 s) &  n/a    \\
    Max episode duration & 880 s ($\pm$ 62 s) &  n/a    \\
    Pretrain+Online time to max & 43044 s ($\pm$ 16106 s) &  n/a \\
    Online time to max & 19088 s ($\pm$ 15011 s) &  n/a \\

    \bottomrule
  \end{tabular}
    \caption{\label{online-timing-table} Wall clock times on the IMDB data set. One episode refers to creating the entire application index set. On average, \textit{Pretrain+Online} reaches its max performance much later in the experiment as it keeps improving while \textit{Online} stops improving early.}
\vspace{-5mm}
\end{table}
\head{Timing.} Next, we analyze time spent in different training phases. Due to small neural network size, pretraining could be comfortably performed within few minutes on a CPU. This includes intermediate evaluations to test accuracy of the model on the set of rule-based demonstrations, and identifying conflicting rules. In table \ref{online-timing-table}, we break down time spent interacting with TensorForce for requesting actions/performing updates, and time spent waiting on the environment and evaluating indexing decisions by running queries. $97.9 \%$ of time was spent waiting on the database to finish creating and removing indices, and only $2.1\%$  was spent on fetching and evaluating actions/queries, and updating the RL model. Pretraining is negligible compared to online evaluation times, so pretraining is desirable if data is available. If LIFT is used for online training, employing pretraining only requires few extra converter methods.

\head{Scalability.}
The indexing problem is complicated by step durations growing with problem scale. Figure \ref{fig:runtimes} shows index creation times for increasing collection sizes. At 100 million documents generated from our synthetic schema, creating an index set for a set of queries can take hours, resulting in weeks of online training. As RL algorithms struggle with data efficiency, we believe these scalability problems will continue to present obstacles for problems such as cluster scheduling. We explore an approach where training is performed on a small data set of 10 million documents. Newly sampled test queries are evaluated on the 100 million document collection without further refinement. Figures \ref{fig:scalability-size}, \ref{fig:scalability-mean-latency}, and \ref{fig:scalability-90-latency} show index size and latencies. All learned strategies created one index per query with query runtimes increasing corresponding to document count, and \textit{Pretrain+Online} performing best. Latency metrics were dominated by a single long-running query with two expensive $\$gt$ expressions which could not be meaningfully accelerated.
While scalability transfer results show some promise, we plan to investigate an approach where a model of the query planner is learned to be able to evaluate indices without needing to run them at full scale.
\begin{table}[h]
  \centering
  \begin{tabular}{lllll}
    \cmidrule{1-5}
      Workload means :  & Mean  & 90th  &  99th  &  Norm. Size (GB) \\
    \midrule
    \textit{Pretrain+Online} & 0.5 s & 1.7 s & 3.5 s & 0.43 \\
    \textit{Online}  & 0.55  s & 2.1 s & 3.5 s & 0.53 \\
    \textit{Default} & 0.94  s & 2.7 s & 3.6 s & 0.03 \\
    \textit{Full} & 0.51 s & 1.5  s & 3.9 s & 1.0 \\
    \textit{Partial} & 0.96 s & 3.4  s & 4.4 s & 0.32 \\
    \textit{Pretrain} & 0.59 s & 2.2  s & 4.1 s & 1.0  \\
    \bottomrule
   
  \end{tabular}
    \caption{\label{variation-table} Performance variation when sampling different query sets per run. \textit{Min}, \textit{90th}, \textit{99th} are referring to average latencies across different query sets.}
   \vspace*{-6mm}
\end{table}

\head{Generalization.} Traditional database benchmarks such as TPC-H use fixed sets of query templates sampling different attribute values at runtime. This is problematic from a deep learning perspective as the number of distinct query shapes (i.e. operator structure) is too small to evaluate generalization to unseen queries. Our synthetic benchmark client does not only sample attribute values on fixed shapes, but also query shapes. We investigate query generalization via our synthetic client by sampling 5 different query sets, and reporting on variation in learning performance. We insert 5 million documents with 15 attributes with varying data types (schema details provided in appendix). Next, $10,000$ queries and rule-based demonstrations are generated using \textit{Full} indexing as the demonstration rule. We did not see improvement when generating more examples, indicating these were sufficient to cover rule behaviour on the synthetic schema. We pretrain on these queries as before, then sample 20 new queries as the test set and perform online training. Table \ref{variation-table} gives an overview on performance variation across query sets. \textit{Pretrain+Online} saves more than $50\%$ space while performing better or comparably across latency metrics.\textit{Partial} saves even more space but fails on improving latency. Values are averaged across different tasks, thus means per task are expected to be different. Importantly, performance of our approach is not an artefact on a specific query set designed for this task but generalizes.


\subsection{Stream task parallelism}
\head{Problem setup.}
Distributed stream processing systems (DSPS) such as Storm \cite{storm}, Heron \cite{Kulkarni2015} or Flink \cite{flink} are widely used in large scale real time processing. To this end, DSPS have to meet strict service level objectives on message latency and throughput. Achieving these objectives requires careful tuning of various scheduling parameters, as processing instances may fail and workloads may vary with sudden spikes. Floratou et al. suggested the notion of self-regulating stream processing with Dhalion \cite{Floratou2017}, a rule-based engine on top of Heron which collect performance metrics, identifies symptoms of performance problems (e.g. instance failure), generates diagnoses and tries to resolve issues by making adjustments (e.g. changing packing plan). We use LIFT to learn to tune task parallelism in Heron using RL. Task parallelism corresponds to the number of physical cores allocated to a specific task in the processing topology. We use the same 3 stage word-count topology as described in Dhalion on a small cluster using 5 machines (1 master, 4 slaves).

\head{Model.}
We again use LIFT to implement state and action models, and to interface Heron's metric collection. For the state, we use a matrix containing CPU and memory usage, and time spent in back-pressure (a special message used by Heron to indicate lack of resources on a task) for all task instances. As actions, the agent outputs (integer) task parallelism values for each component in the topology. The reward is a linear combination of normalized message latencies square roots (to smooth outliers), throughput, and the fraction of available instances used.
\begin{figure}[t] 
\centering
\includegraphics[scale=.55]{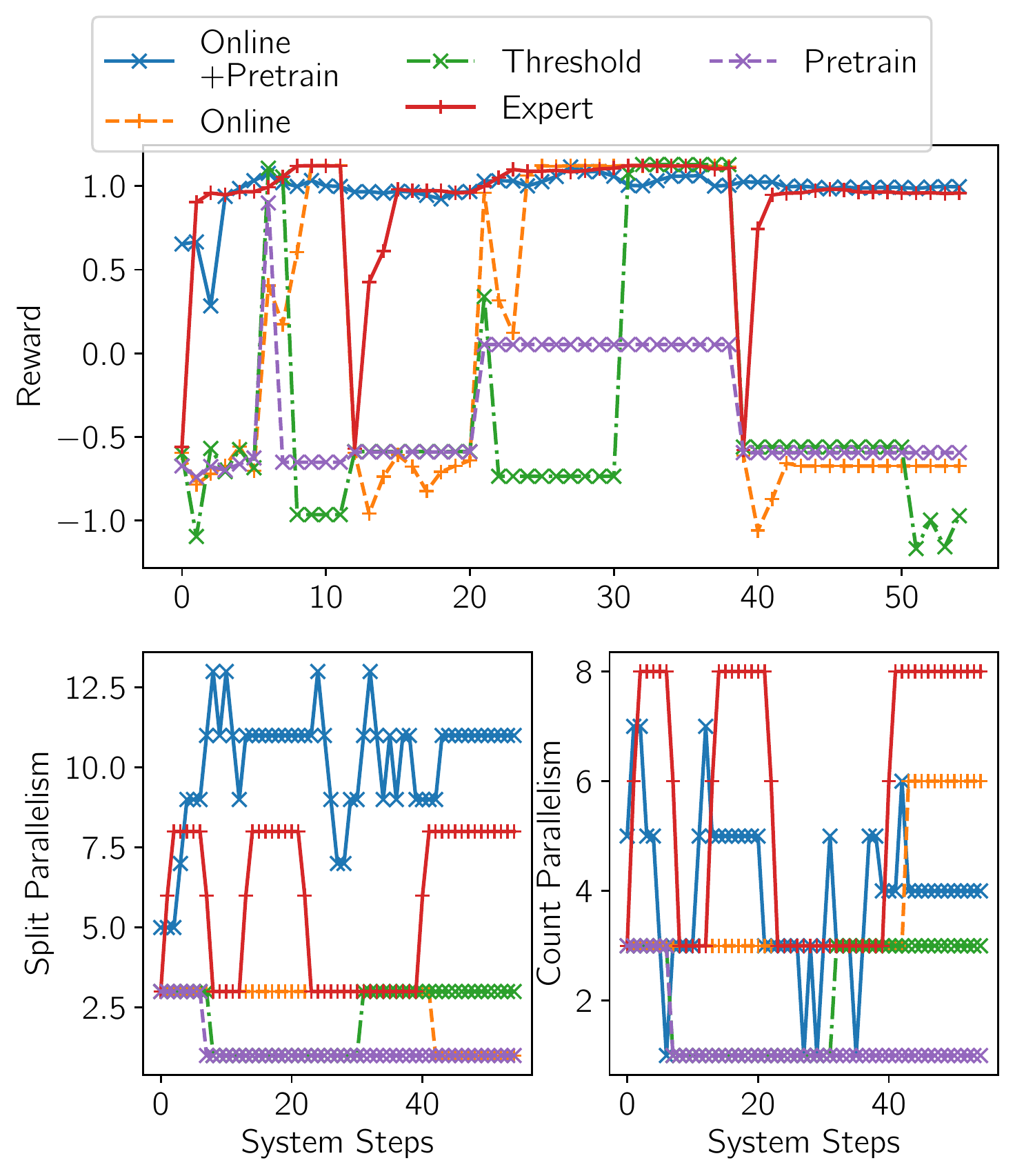}
\caption{Top: Rewards through varying workload. Bottom: Task parallelisms for splitter and count bolts.}
\label{fig:heron-rewards}
\vspace{-7mm}
\end{figure} 

\head{Results.}  Collecting data for the stream processing task is difficult as each step requires multiple minutes of collecting metrics so performance can stabilize after changes, and updating the topology by creating a new packing plan and deploying it. Due to an outstanding issue in the scheduler, we did not manage to run Dhalion itself. We could also not easily port its parallelism rule to LIFT because not all used metrics were exposed via the Heron tracker API. For the purpose of this experiment, we hence collected demonstration data from a simple threshold rule. The aim of this case study is hence not to prove superiority over Dhalion, but evaluate if rule-based demonstrations can help RL in dynamic workload environments. We train and evaluate dynamic behavior by randomly sampling different workload changes such as load moving up and down periodically, or changing from low to high/high to low. Figure \ref{fig:heron-rewards} shows results by comparing average reward over the duration of the evaluation which presented the controller with all possible workloads in deterministic order. Each step corresponds to about 2-4 minutes real time to receive metrics and implement changes. 

We defined an \textit{Expert} configuration which had predetermined good configurations for each workload change. The bottom row shows how parallelism settings for both bolts are adjusted by the different strategies over time. The \textit{Expert} systematically alternates between two configurations for each component, incurring temporary low rewards upon changes. The \textit{Pretrain+Online} agent managed to avoid temporary reward loss by anticipating workload changes, and by always keeping split parallelism high as to have enough capacity for changes, thus outperforming the pre-tuned \textit{Expert} configurations slightly ($3\%$. This 'anticipation effect' is a consequence of the agent observing regularities in how workloads change. \textit{Online} failed to adopt an effective strategy within the same training time (1.5 days). Other methods performed worse although the threshold rule-based model could have been improved by manually fitting thresholds to workload changes (thus being closer to \textit{Expert}). 

\begin{figure}[t]
\centering
\begin{subfigure}[b]{0.4\textwidth}
   \includegraphics[scale=0.35,width=1\linewidth]{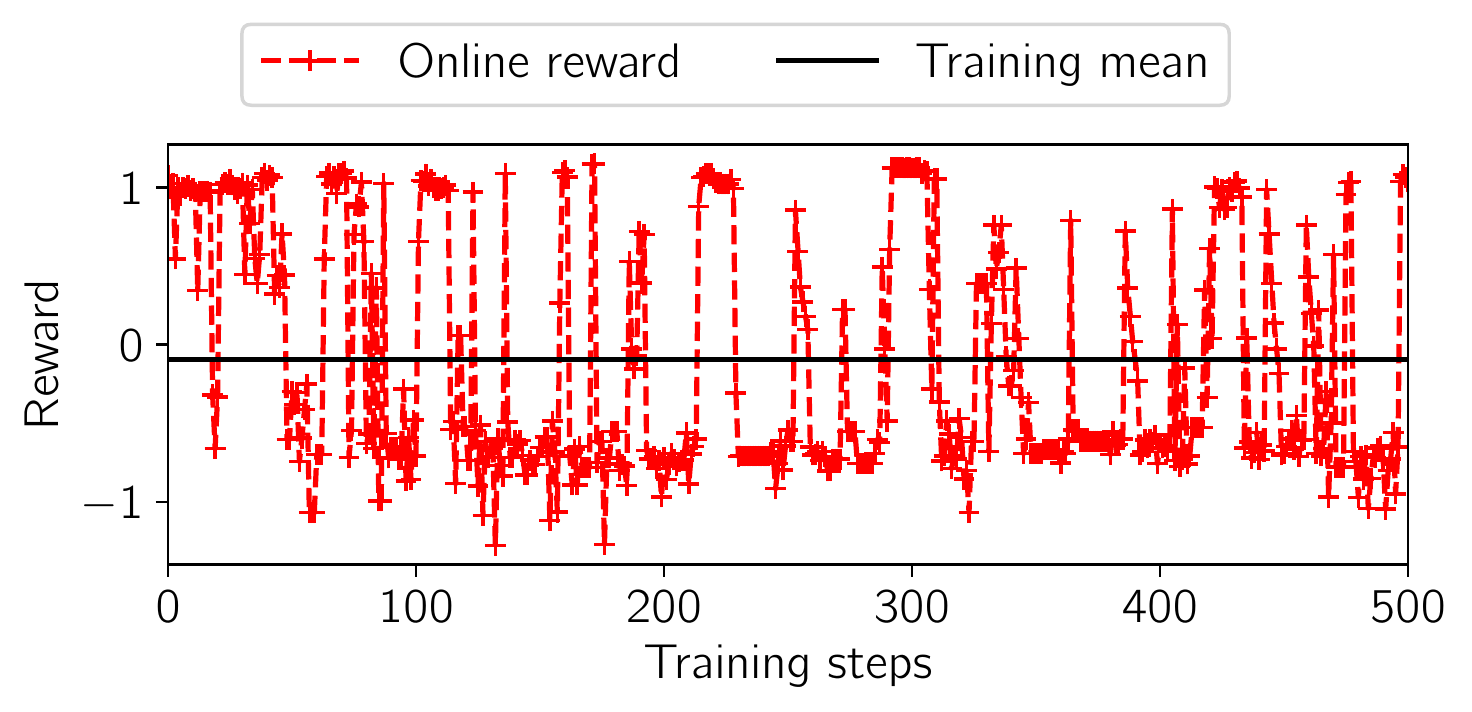}
\end{subfigure}

\begin{subfigure}[b]{0.4\textwidth}
   \includegraphics[scale=0.35,width=1\linewidth]{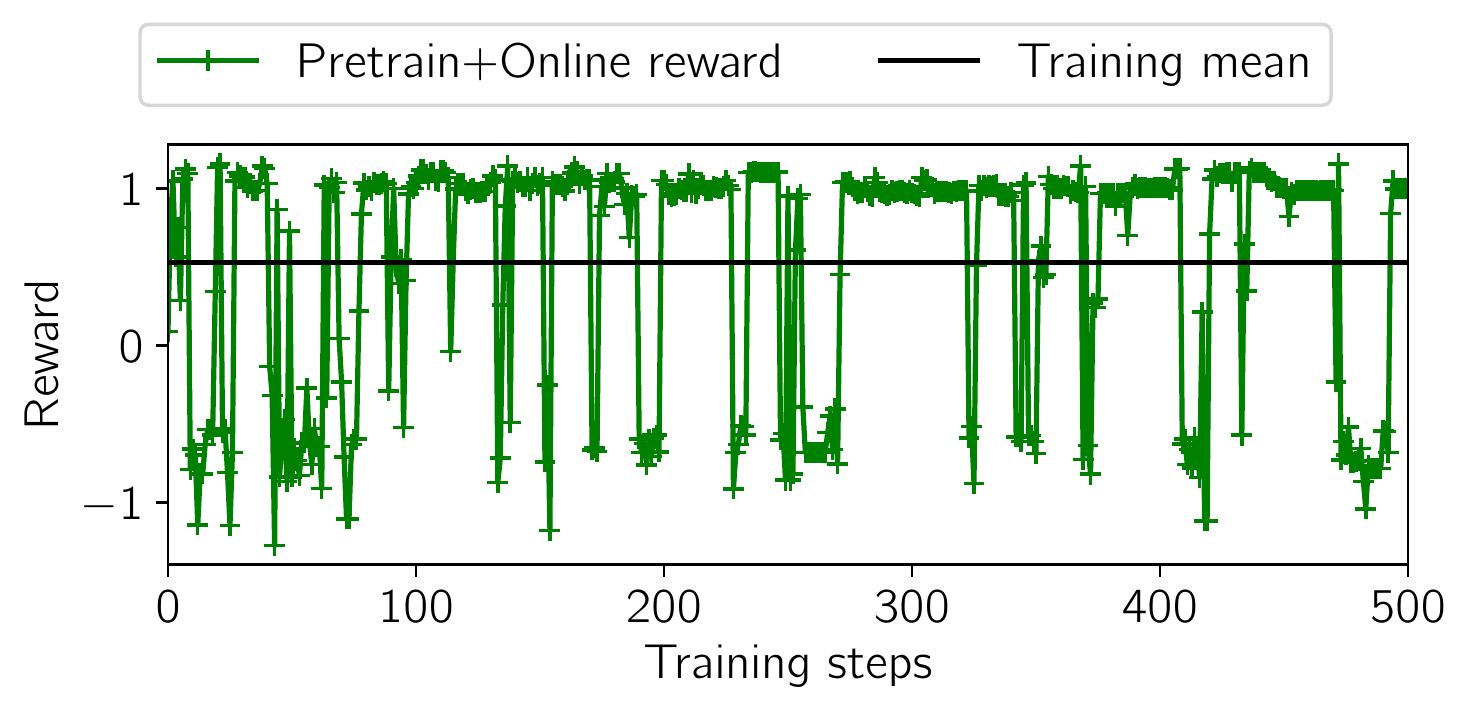}
\end{subfigure}
\caption{Heron training rewards.\label{fig:heron-training-rewards}}
\vspace{-6mm}
\end{figure}

We provide further analysis by comparing training rewards with and without pretraining in Figure \ref{fig:heron-training-rewards}. \textit{Online} without pretraining could on average not recover good configurations, thus most of the time being at a low reward, and only occasionally seeing high rewards when workloads matched its configuration. In contrast, \textit{Pretrain+Online} achieved much higher mean rewards as after around 100 episodes of training, it began to quickly recover from workload changes to reach (normalized) high reward regions again. Our experiments show the combination of pretraining and online refinement can produce effective results on dynamic workloads. A key question is if workloads in practice exhibit irregularities which are difficult to address through manual tuning. We suspect the advantage of RL will increase for larger topologies with many different bolt types on heterogeneous resources.

\subsection{Discussion}
\head{Limitations.}
Our results indicate the potential of imperfect demonstrations when applying RL to data management tasks, improving latency metrics by up to $70\%$ in the IMDB case study against several baselines, and outperforming the expert configuration in Heron. Our experiments did not include some subtasks which prolong training times. For example, in the indexing task, we omitted considerations for indexing shards and replica sets. As RL applications in data management move from simulation to real world, they will incrementally cover additional subtasks. We also showed the difficulty of tackling tasks where step times increase with scale. Here, mechanisms such as pretraining and training on partial tasks provide a promising direction to eventually apply RL at data center scale.

\head{Learning.}
The algorithmic limitations of current RL algorithms continue to present significant limitations in real world applications. Learning from scratch can take infeasibly long and may also be unreliable due to the stochastic nature of training. Further, learning a task once and deploying the resulting model in different contexts is unlikely to succeed due to the sensitivity of RL algorithms \cite{Kansky2017}. We rely on online refinement following the pretraining procedure which incurs only little overhead after initial implementation. The aim of our experiments was not to demonstrate the best way to e.g. perform workload management in stream processing. Recent work has illustrated how neural networks struggle with forecasting spiky workloads \cite{Ma2018}. We focused on evaluating if pretraining can help the notoriously difficult application of RL to data management tasks. In summary, DRL remains a promising direction, as even results in this exploratory phase show the ability to learn complex behaviour which normally requires manual solution design.


\section{Related work}
\head{RL in data management.}
Early work in exploring RL in computer systems can be found in routing (Q-routing) and protocol optimization \cite{BoyanLittman1994, KumarMiikkulainen1997,KumarMiikkulainen1999}. Subsequent research has covered a diverse range of domains such as cloud workload allocation, cluster scheduling, networking, or bitrate selection \cite{TesauroDasChanEtAl2007, DutreilhKirgizovMelekhovaEtAl2011, Mao2016, Mao2017, Valadarsky2017a}. Many of these works have outperformed existing approaches in simulation, but did not translate into real world deployments due to the difficulties discussed elsewhere in this paper. Notably, the idea of using neural networks in combination with RL in systems can be found as early as 2006 in Tesauro et al.'s work on server resource allocation~\cite{TesauroJongDasEtAl2006}. The authors incorporated pre-existing knowledge by initially bootstrapping control from a rule, whereas we use offline demonstrations to reduce training times. RL for combinatorial selection has also found application in tuning compression of neural networks for mobile devices \cite{Liu2018}. Mirhoseini et al. demonstrated how to use attention-based andwe p hierarchical methods known from neural machine translation to effectively perform TensorFlow device placements \cite{mirhoseini2017device, hierarchical2018}. The difference to our work is that each graph step only takes few seconds so online training can be performed more effectively. Sharma et al. used RL to learn single-key indices in relational databases, and simplified the problem by manually constructing features such as selectivity \cite{Sharma2018}. 

\head{Adaptive indexing.}
A large body of work exists on indexing strategies which are widely used in practice. Offline indexing is performed using the design tuning tools provided by commercial database products which require database administrators to manually interact with the tool, and make ultimate design decisions \cite{dbtuningsqlserver2005, Chaudhuri1998, Dageville2004}. Online indexing addresses this limitation by making decisions based on continuous online monitoring \cite{Schnaitter2006}. Adaptive (or holistic \cite{Petraki2015}) indexing (e.g. in columnar databases) enable even faster reaction to workload changes by building and refining indices via lightweight incremental modifications \cite{Graefe2010, Idreos2011, Halim2012}. A similar depth of work in indexing is not available for document databases, although many techniques are likely transferable \cite{Qader2018}. Commercial MongoDB services sometimes offer index recommendations based on longer term workload patterns (\cite{mlab2018}).

\head{ML in databases.} Recently, machine learning approaches have been explored in data management. Pavlo et al. proposed the idea of a self-driving database with initial focus on employing ML techniques for workload forecasting \cite{Pavlo2017}. In subsequent work, these forecasts were evaluated on their ability to help create SQL indices \cite{Ma2018}. Their work in particular found that neural networks were not as effective in capturing spiky loads as traditional time series techniques. OtterTune \cite{VanAken2017} automatically determines relevant database parameters to create end-to-end tuning pipelines \cite{Golovin2017}. BO is not easily applicable in problems like index selection or generally combinatorial problems as it requires a similarity function (Kernel) to interpolate a smooth objective function between data points. Defining a custom Kernel between databases is difficult because semantically similar indices can perform vastly different on a workload. Kraska et al. explored representing the index data structure itself as a neural network wit the aim to learn to match data distributions and access patterns \cite{Kraska2017}. Bailis et al. subsequently argued that well tuned cuckoo hashing could still outperform these learned indices \cite{bailis2018learning}. We similarly argue that deep RL techniques are in an exploratory phase and cannot yet replace well established tuning methods.

\head{Learning from demonstrations.}
Learning from expert demonstration is a well studied notion in RL. DAGGER (for Dataset Aggregation) is a popular approach for imitation learning which requires an expert to continuously provide new input \cite{Ross2010}. While this is not directly compatible with learning from traces, we are considering future additions to LIFT where a human may interactively provide demonstrations between trials to further accelerate training. Other familiar approaches include behavioural cloning, recently also in the context of generative adversarial models \cite{Wang2017, Ho2016}. Snorkel is a system to help generate weakly supervised data via labeling functions which is conceptually similar to our rule-based demonstrations \cite{Ratner2017}. In this work, we relied on DQfD as a conceptually simple extension to Deep Q-learning \cite{Hester17}. Its main advantage is that the large margin function gives a simple way of assigning low or high confidence to demonstrations via single tunable parameter, thus in practice also allowing the use of imperfect demonstrations. Gao et al. recently suggested employing a unified objective which incorporates imperfect demonstrations naturally by accounting for uncertainty using an entropy approach \cite{Gao2018}. 

\section{Conclusion}
In this paper, we discuss long evaluation times, algorithmic instability, and lack of widely available software as key obstacles to the applicability of RL in data management tasks. To help address these issues, we introduce LIFT, the first end-to-end software stack for applying RL to data management. As part of LIFT, we also introduce TensorForce, a practical deep reinforcement learning library providing a declarative API to common RL algorithms. The key idea of LIFT is to help developers leveraging existing knowledge from trace data, rules or any other form of demonstrations to guide model creation. We demonstrate the practical potential of LIFT in two proof-of-concept case studies. If online-only training takes impractically long, our results show that pretraining can significantly improve final results.
\bibliographystyle{ACM-Reference-Format}

\begin{thebibliography}{76}


\ifx \showCODEN    \undefined \def \showCODEN     #1{\unskip}     \fi
\ifx \showDOI      \undefined \def \showDOI       #1{#1}\fi
\ifx \showISBNx    \undefined \def \showISBNx     #1{\unskip}     \fi
\ifx \showISBNxiii \undefined \def \showISBNxiii  #1{\unskip}     \fi
\ifx \showISSN     \undefined \def \showISSN      #1{\unskip}     \fi
\ifx \showLCCN     \undefined \def \showLCCN      #1{\unskip}     \fi
\ifx \shownote     \undefined \def \shownote      #1{#1}          \fi
\ifx \showarticletitle \undefined \def \showarticletitle #1{#1}   \fi
\ifx \showURL      \undefined \def \showURL       {\relax}        \fi
\providecommand\bibfield[2]{#2}
\providecommand\bibinfo[2]{#2}
\providecommand\natexlab[1]{#1}
\providecommand\showeprint[2][]{arXiv:#2}

\bibitem[\protect\citeauthoryear{Abadi, Agarwal, Barham, Brevdo, Chen, Citro,
  Corrado, Davis, Dean, Devin, et~al\mbox{.}}{Abadi et~al\mbox{.}}{2015}]%
        {AbadiAgarwalBarhamEtAl2016}
\bibfield{author}{\bibinfo{person}{Mart{\i}n Abadi}, \bibinfo{person}{Ashish
  Agarwal}, \bibinfo{person}{Paul Barham}, \bibinfo{person}{Eugene Brevdo},
  \bibinfo{person}{Zhifeng Chen}, \bibinfo{person}{Craig Citro},
  \bibinfo{person}{Greg~S Corrado}, \bibinfo{person}{Andy Davis},
  \bibinfo{person}{Jeffrey Dean}, \bibinfo{person}{Matthieu Devin},
  {et~al\mbox{.}}} \bibinfo{year}{2015}\natexlab{}.
\newblock \showarticletitle{TensorFlow: Large-Scale Machine Learning on
  Heterogeneous Distributed Systems}.
\newblock \bibinfo{journal}{\emph{arXiv preprint arXiv:1603.04467}}
  (\bibinfo{year}{2015}).
\newblock


\bibitem[\protect\citeauthoryear{Abadi, Barham, Chen, Chen, Davis, Dean, Devin,
  Ghemawat, Irving, Isard, et~al\mbox{.}}{Abadi et~al\mbox{.}}{2016}]%
        {abadi2016tensorflow}
\bibfield{author}{\bibinfo{person}{Mart{\'\i}n Abadi}, \bibinfo{person}{Paul
  Barham}, \bibinfo{person}{Jianmin Chen}, \bibinfo{person}{Zhifeng Chen},
  \bibinfo{person}{Andy Davis}, \bibinfo{person}{Jeffrey Dean},
  \bibinfo{person}{Matthieu Devin}, \bibinfo{person}{Sanjay Ghemawat},
  \bibinfo{person}{Geoffrey Irving}, \bibinfo{person}{Michael Isard},
  {et~al\mbox{.}}} \bibinfo{year}{2016}\natexlab{}.
\newblock \showarticletitle{TensorFlow: A System for Large-Scale Machine
  Learning.}. In \bibinfo{booktitle}{\emph{OSDI}}, Vol.~\bibinfo{volume}{16}.
  \bibinfo{pages}{265--283}.
\newblock


\bibitem[\protect\citeauthoryear{Agrawal, Chaudhuri, Kollar, Marathe,
  Narasayya, and Syamala}{Agrawal et~al\mbox{.}}{2004}]%
        {dbtuningsqlserver2005}
\bibfield{author}{\bibinfo{person}{Sanjay Agrawal}, \bibinfo{person}{Surajit
  Chaudhuri}, \bibinfo{person}{Lubor Kollar}, \bibinfo{person}{Arun Marathe},
  \bibinfo{person}{Vivek Narasayya}, {and} \bibinfo{person}{Manoj Syamala}.}
  \bibinfo{year}{2004}\natexlab{}.
\newblock \showarticletitle{Database Tuning Advisor for Microsoft SQL Server
  2005}. In \bibinfo{booktitle}{\emph{VLDB}}. \bibinfo{publisher}{Very Large
  Data Bases Endowment Inc.}
\newblock
\urldef\tempurl%
\url{https://www.microsoft.com/en-us/research/publication/database-tuning-advisor-for-microsoft-sql-server-2005/}
\showURL{%
\tempurl}


\bibitem[\protect\citeauthoryear{{Amazon Inc.}}{{Amazon Inc.}}{2018}]%
        {dynamodb}
\bibfield{author}{\bibinfo{person}{{Amazon Inc.}}}
  \bibinfo{year}{2018}\natexlab{}.
\newblock \bibinfo{title}{Amazon DynamoDB}.
\newblock \bibinfo{howpublished}{website}.   (\bibinfo{year}{2018}).
\newblock
\urldef\tempurl%
\url{https://aws.amazon.com/dynamodb/}
\showURL{%
\tempurl}


\bibitem[\protect\citeauthoryear{{Apache Foundation}}{{Apache
  Foundation}}{2018}]%
        {flink}
\bibfield{author}{\bibinfo{person}{{Apache Foundation}}.}
  \bibinfo{year}{2018}\natexlab{}.
\newblock \bibinfo{title}{Apache Flink}.
\newblock \bibinfo{howpublished}{website}.   (\bibinfo{year}{2018}).
\newblock
\urldef\tempurl%
\url{https://flink.apache.org}
\showURL{%
\tempurl}


\bibitem[\protect\citeauthoryear{Arulkumaran, Deisenroth, Brundage, and
  Bharath}{Arulkumaran et~al\mbox{.}}{2017}]%
        {arulkumaran2017brief}
\bibfield{author}{\bibinfo{person}{Kai Arulkumaran},
  \bibinfo{person}{Marc~Peter Deisenroth}, \bibinfo{person}{Miles Brundage},
  {and} \bibinfo{person}{Anil~Anthony Bharath}.}
  \bibinfo{year}{2017}\natexlab{}.
\newblock \showarticletitle{A Brief Survey of Deep Reinforcement Learning}.
\newblock \bibinfo{journal}{\emph{arXiv preprint arXiv:1708.05866}}
  (\bibinfo{year}{2017}).
\newblock


\bibitem[\protect\citeauthoryear{Bellemare, Naddaf, Veness, and
  Bowling}{Bellemare et~al\mbox{.}}{2013}]%
        {Bellemare2013}
\bibfield{author}{\bibinfo{person}{Marc~G Bellemare}, \bibinfo{person}{Yavar
  Naddaf}, \bibinfo{person}{Joel Veness}, {and} \bibinfo{person}{Michael
  Bowling}.} \bibinfo{year}{2013}\natexlab{}.
\newblock \showarticletitle{The Arcade Learning Environment: An evaluation
  platform for general agents.}
\newblock \bibinfo{journal}{\emph{J. Artif. Intell. Res.(JAIR)}}
  \bibinfo{volume}{47} (\bibinfo{year}{2013}), \bibinfo{pages}{253--279}.
\newblock


\bibitem[\protect\citeauthoryear{Boyan and Littman}{Boyan and Littman}{1994}]%
        {BoyanLittman1994}
\bibfield{author}{\bibinfo{person}{Justin~A Boyan} {and}
  \bibinfo{person}{Michael~L Littman}.} \bibinfo{year}{1994}\natexlab{}.
\newblock \showarticletitle{Packet routing in dynamically changing networks: A
  reinforcement learning approach}.
\newblock \bibinfo{journal}{\emph{Advances in neural information processing
  systems}} (\bibinfo{year}{1994}), \bibinfo{pages}{671--671}.
\newblock


\bibitem[\protect\citeauthoryear{Brockman, Cheung, Pettersson, Schneider,
  Schulman, Tang, and Zaremba}{Brockman et~al\mbox{.}}{2016}]%
        {brockman2016openai}
\bibfield{author}{\bibinfo{person}{Greg Brockman}, \bibinfo{person}{Vicki
  Cheung}, \bibinfo{person}{Ludwig Pettersson}, \bibinfo{person}{Jonas
  Schneider}, \bibinfo{person}{John Schulman}, \bibinfo{person}{Jie Tang},
  {and} \bibinfo{person}{Wojciech Zaremba}.} \bibinfo{year}{2016}\natexlab{}.
\newblock \showarticletitle{OpenAI gym}.
\newblock \bibinfo{journal}{\emph{arXiv preprint arXiv:1606.01540}}
  (\bibinfo{year}{2016}).
\newblock


\bibitem[\protect\citeauthoryear{Caspi, Leibovich, and Novik}{Caspi
  et~al\mbox{.}}{2017}]%
        {nervanacoach}
\bibfield{author}{\bibinfo{person}{Itai Caspi}, \bibinfo{person}{Gal
  Leibovich}, {and} \bibinfo{person}{Gal Novik}.}
  \bibinfo{year}{2017}\natexlab{}.
\newblock \bibinfo{title}{Reinforcement Learning Coach}.
\newblock   (\bibinfo{date}{Dec.} \bibinfo{year}{2017}).
\newblock
\urldef\tempurl%
\url{https://doi.org/10.5281/zenodo.1134899}
\showDOI{\tempurl}


\bibitem[\protect\citeauthoryear{Chaudhuri and Narasayya}{Chaudhuri and
  Narasayya}{1998}]%
        {Chaudhuri1998}
\bibfield{author}{\bibinfo{person}{Surajit Chaudhuri} {and}
  \bibinfo{person}{Vivek Narasayya}.} \bibinfo{year}{1998}\natexlab{}.
\newblock \showarticletitle{AutoAdmin \&Ldquo;What-if\&Rdquo; Index Analysis
  Utility}.
\newblock \bibinfo{journal}{\emph{SIGMOD Rec.}} \bibinfo{volume}{27},
  \bibinfo{number}{2} (\bibinfo{date}{June} \bibinfo{year}{1998}),
  \bibinfo{pages}{367--378}.
\newblock
\showISSN{0163-5808}
\urldef\tempurl%
\url{https://doi.org/10.1145/276305.276337}
\showDOI{\tempurl}


\bibitem[\protect\citeauthoryear{Chen, Li, Li, Lin, Wang, Wang, Xiao, Xu,
  Zhang, and Zhang}{Chen et~al\mbox{.}}{2015}]%
        {chen2015mxnet}
\bibfield{author}{\bibinfo{person}{Tianqi Chen}, \bibinfo{person}{Mu Li},
  \bibinfo{person}{Yutian Li}, \bibinfo{person}{Min Lin},
  \bibinfo{person}{Naiyan Wang}, \bibinfo{person}{Minjie Wang},
  \bibinfo{person}{Tianjun Xiao}, \bibinfo{person}{Bing Xu},
  \bibinfo{person}{Chiyuan Zhang}, {and} \bibinfo{person}{Zheng Zhang}.}
  \bibinfo{year}{2015}\natexlab{}.
\newblock \showarticletitle{Mxnet: A flexible and efficient machine learning
  library for heterogeneous distributed systems}.
\newblock \bibinfo{journal}{\emph{arXiv preprint arXiv:1512.01274}}
  (\bibinfo{year}{2015}).
\newblock


\bibitem[\protect\citeauthoryear{Cooper, Silberstein, Tam, Ramakrishnan, and
  Sears}{Cooper et~al\mbox{.}}{2010}]%
        {CooperSilbersteinTamEtAl2010}
\bibfield{author}{\bibinfo{person}{Brian~F. Cooper}, \bibinfo{person}{Adam
  Silberstein}, \bibinfo{person}{Erwin Tam}, \bibinfo{person}{Raghu
  Ramakrishnan}, {and} \bibinfo{person}{Russell Sears}.}
  \bibinfo{year}{2010}\natexlab{}.
\newblock \showarticletitle{Benchmarking Cloud Serving Systems with YCSB}. In
  \bibinfo{booktitle}{\emph{Proceedings of the 1st ACM Symposium on Cloud
  Computing}} \emph{(\bibinfo{series}{SoCC '10})}. \bibinfo{publisher}{ACM},
  \bibinfo{address}{New York, NY, USA}, \bibinfo{pages}{143--154}.
\newblock
\showISBNx{978-1-4503-0036-0}
\urldef\tempurl%
\url{https://doi.org/10.1145/1807128.1807152}
\showDOI{\tempurl}


\bibitem[\protect\citeauthoryear{Dageville, Das, Dias, Yagoub, Zait, and
  Ziauddin}{Dageville et~al\mbox{.}}{2004}]%
        {Dageville2004}
\bibfield{author}{\bibinfo{person}{Benoit Dageville}, \bibinfo{person}{Dinesh
  Das}, \bibinfo{person}{Karl Dias}, \bibinfo{person}{Khaled Yagoub},
  \bibinfo{person}{Mohamed Zait}, {and} \bibinfo{person}{Mohamed Ziauddin}.}
  \bibinfo{year}{2004}\natexlab{}.
\newblock \showarticletitle{Automatic SQL Tuning in Oracle 10G}. In
  \bibinfo{booktitle}{\emph{Proceedings of the Thirtieth International
  Conference on Very Large Data Bases - Volume 30}}
  \emph{(\bibinfo{series}{VLDB '04})}. \bibinfo{publisher}{VLDB Endowment},
  \bibinfo{pages}{1098--1109}.
\newblock
\showISBNx{0-12-088469-0}
\urldef\tempurl%
\url{http://dl.acm.org/citation.cfm?id=1316689.1316784}
\showURL{%
\tempurl}


\bibitem[\protect\citeauthoryear{Dutreilh, Kirgizov, Melekhova, Malenfant,
  Rivierre, and Truck}{Dutreilh et~al\mbox{.}}{2011}]%
        {DutreilhKirgizovMelekhovaEtAl2011}
\bibfield{author}{\bibinfo{person}{Xavier Dutreilh}, \bibinfo{person}{Sergey
  Kirgizov}, \bibinfo{person}{Olga Melekhova}, \bibinfo{person}{Jacques
  Malenfant}, \bibinfo{person}{Nicolas Rivierre}, {and} \bibinfo{person}{Isis
  Truck}.} \bibinfo{year}{2011}\natexlab{}.
\newblock \showarticletitle{Using reinforcement learning for autonomic resource
  allocation in clouds: towards a fully automated workflow}. In
  \bibinfo{booktitle}{\emph{{ICAS} 2011, The Seventh International Conference
  on Autonomic and Autonomous Systems}}. \bibinfo{pages}{67{\textendash}74}.
\newblock


\bibitem[\protect\citeauthoryear{Espeholt, Soyer, Munos, Simonyan, Mnih, Ward,
  Doron, Firoiu, Harley, Dunning, Legg, and Kavukcuoglu}{Espeholt
  et~al\mbox{.}}{[n. d.]}]%
        {Espeholt2018}
\bibfield{author}{\bibinfo{person}{Lasse Espeholt}, \bibinfo{person}{Hubert
  Soyer}, \bibinfo{person}{Remi Munos}, \bibinfo{person}{Karen Simonyan},
  \bibinfo{person}{Volodymir Mnih}, \bibinfo{person}{Tom Ward},
  \bibinfo{person}{Yotam Doron}, \bibinfo{person}{Vlad Firoiu},
  \bibinfo{person}{Tim Harley}, \bibinfo{person}{Iain Dunning},
  \bibinfo{person}{Shane Legg}, {and} \bibinfo{person}{Koray Kavukcuoglu}.}
  \bibinfo{year}{[n. d.]}\natexlab{}.
\newblock \showarticletitle{IMPALA: Scalable Distributed Deep-RL with
  Importance Weighted Actor-Learner Architectures}.
\newblock  (\bibinfo{year}{[n. d.]}).
\newblock
\showeprint[arXiv]{cs.LG/1802.01561v2}


\bibitem[\protect\citeauthoryear{Floratou, Agrawal, Graham, Rao, and
  Ramasamy}{Floratou et~al\mbox{.}}{2017}]%
        {Floratou2017}
\bibfield{author}{\bibinfo{person}{Avrilia Floratou}, \bibinfo{person}{Ashvin
  Agrawal}, \bibinfo{person}{Bill Graham}, \bibinfo{person}{Sriram Rao}, {and}
  \bibinfo{person}{Karthik Ramasamy}.} \bibinfo{year}{2017}\natexlab{}.
\newblock \showarticletitle{Dhalion: self-regulating stream processing in
  heron}.
\newblock \bibinfo{journal}{\emph{Proceedings of the VLDB Endowment}}
  \bibinfo{volume}{10}, \bibinfo{number}{12} (\bibinfo{year}{2017}),
  \bibinfo{pages}{1825--1836}.
\newblock


\bibitem[\protect\citeauthoryear{Gao, Huazhe, Xu, Lin, Yu, Levine, and
  Darrell}{Gao et~al\mbox{.}}{[n. d.]}]%
        {Gao2018}
\bibfield{author}{\bibinfo{person}{Yang Gao}, \bibinfo{person}{Huazhe},
  \bibinfo{person}{Xu}, \bibinfo{person}{Ji Lin}, \bibinfo{person}{Fisher Yu},
  \bibinfo{person}{Sergey Levine}, {and} \bibinfo{person}{Trevor Darrell}.}
  \bibinfo{year}{[n. d.]}\natexlab{}.
\newblock \showarticletitle{Reinforcement Learning from Imperfect
  Demonstrations}.
\newblock  (\bibinfo{year}{[n. d.]}).
\newblock
\showeprint[arXiv]{cs.AI/1802.05313v1}


\bibitem[\protect\citeauthoryear{Golovin, Solnik, Moitra, Kochanski, Karro, and
  Sculley}{Golovin et~al\mbox{.}}{2017}]%
        {Golovin2017}
\bibfield{author}{\bibinfo{person}{Daniel Golovin}, \bibinfo{person}{Benjamin
  Solnik}, \bibinfo{person}{Subhodeep Moitra}, \bibinfo{person}{Greg
  Kochanski}, \bibinfo{person}{John Karro}, {and} \bibinfo{person}{D Sculley}.}
  \bibinfo{year}{2017}\natexlab{}.
\newblock \showarticletitle{Google Vizier: A Service for Black-Box
  Optimization}. In \bibinfo{booktitle}{\emph{Proceedings of the 23rd ACM
  SIGKDD International Conference on Knowledge Discovery and Data Mining}}.
  ACM, \bibinfo{pages}{1487--1495}.
\newblock


\bibitem[\protect\citeauthoryear{{Google Inc.}}{{Google Inc.}}{2018}]%
        {googlecloudstore}
\bibfield{author}{\bibinfo{person}{{Google Inc.}}}
  \bibinfo{year}{2018}\natexlab{}.
\newblock \bibinfo{title}{Google Cloud Datastore}.
\newblock \bibinfo{howpublished}{website}.   (\bibinfo{year}{2018}).
\newblock
\urldef\tempurl%
\url{https://cloud.google.com/datastore/}
\showURL{%
\tempurl}


\bibitem[\protect\citeauthoryear{Graefe and Kuno}{Graefe and Kuno}{2010}]%
        {Graefe2010}
\bibfield{author}{\bibinfo{person}{Goetz Graefe} {and} \bibinfo{person}{Harumi
  Kuno}.} \bibinfo{year}{2010}\natexlab{}.
\newblock \showarticletitle{Self-selecting, Self-tuning, Incrementally
  Optimized Indexes}. In \bibinfo{booktitle}{\emph{Proceedings of the 13th
  International Conference on Extending Database Technology}}
  \emph{(\bibinfo{series}{EDBT '10})}. \bibinfo{publisher}{ACM},
  \bibinfo{address}{New York, NY, USA}, \bibinfo{pages}{371--381}.
\newblock
\showISBNx{978-1-60558-945-9}
\urldef\tempurl%
\url{https://doi.org/10.1145/1739041.1739087}
\showDOI{\tempurl}


\bibitem[\protect\citeauthoryear{Gu, Lillicrap, Sutskever, and Levine}{Gu
  et~al\mbox{.}}{2016}]%
        {GuLillicrapSutskeverEtAl2016}
\bibfield{author}{\bibinfo{person}{Shixiang Gu}, \bibinfo{person}{Timothy
  Lillicrap}, \bibinfo{person}{Ilya Sutskever}, {and} \bibinfo{person}{Sergey
  Levine}.} \bibinfo{year}{2016}\natexlab{}.
\newblock \showarticletitle{Continuous Deep Q-Learning with Model-based
  Acceleration}.
\newblock  (\bibinfo{date}{March} \bibinfo{year}{2016}).
\newblock
\showeprint{1603.00748}


\bibitem[\protect\citeauthoryear{Haarnoja, Tang, Abbeel, and Levine}{Haarnoja
  et~al\mbox{.}}{2017}]%
        {Haarnoja2017}
\bibfield{author}{\bibinfo{person}{Tuomas Haarnoja}, \bibinfo{person}{Haoran
  Tang}, \bibinfo{person}{Pieter Abbeel}, {and} \bibinfo{person}{Sergey
  Levine}.} \bibinfo{year}{2017}\natexlab{}.
\newblock \showarticletitle{Reinforcement learning with deep energy-based
  policies}.
\newblock \bibinfo{journal}{\emph{arXiv preprint arXiv:1702.08165}}
  (\bibinfo{year}{2017}).
\newblock


\bibitem[\protect\citeauthoryear{Halim, Idreos, Karras, and Yap}{Halim
  et~al\mbox{.}}{2012}]%
        {Halim2012}
\bibfield{author}{\bibinfo{person}{Felix Halim}, \bibinfo{person}{Stratos
  Idreos}, \bibinfo{person}{Panagiotis Karras}, {and} \bibinfo{person}{Roland
  H.~C. Yap}.} \bibinfo{year}{2012}\natexlab{}.
\newblock \showarticletitle{Stochastic Database Cracking: Towards Robust
  Adaptive Indexing in Main-memory Column-stores}.
\newblock \bibinfo{journal}{\emph{Proc. VLDB Endow.}} \bibinfo{volume}{5},
  \bibinfo{number}{6} (\bibinfo{date}{Feb.} \bibinfo{year}{2012}),
  \bibinfo{pages}{502--513}.
\newblock
\showISSN{2150-8097}
\urldef\tempurl%
\url{https://doi.org/10.14778/2168651.2168652}
\showDOI{\tempurl}


\bibitem[\protect\citeauthoryear{Hein, Depeweg, Tokic, Udluft, Hentschel,
  Runkler, and Sterzing}{Hein et~al\mbox{.}}{2017}]%
        {Hein2017a}
\bibfield{author}{\bibinfo{person}{Daniel Hein}, \bibinfo{person}{Stefan
  Depeweg}, \bibinfo{person}{Michel Tokic}, \bibinfo{person}{Steffen Udluft},
  \bibinfo{person}{Alexander Hentschel}, \bibinfo{person}{Thomas~A Runkler},
  {and} \bibinfo{person}{Volkmar Sterzing}.} \bibinfo{year}{2017}\natexlab{}.
\newblock \showarticletitle{A Benchmark Environment Motivated by Industrial
  Control Problems}.
\newblock \bibinfo{journal}{\emph{arXiv preprint arXiv:1709.09480}}
  (\bibinfo{year}{2017}).
\newblock


\bibitem[\protect\citeauthoryear{Henderson, Islam, Bachman, Pineau, Precup, and
  Meger}{Henderson et~al\mbox{.}}{2017}]%
        {Henderson2017}
\bibfield{author}{\bibinfo{person}{Peter Henderson}, \bibinfo{person}{Riashat
  Islam}, \bibinfo{person}{Philip Bachman}, \bibinfo{person}{Joelle Pineau},
  \bibinfo{person}{Doina Precup}, {and} \bibinfo{person}{David Meger}.}
  \bibinfo{year}{2017}\natexlab{}.
\newblock \showarticletitle{Deep reinforcement learning that matters}.
\newblock \bibinfo{journal}{\emph{arXiv preprint arXiv:1709.06560}}
  (\bibinfo{year}{2017}).
\newblock


\bibitem[\protect\citeauthoryear{Hester, Vecerik, Pietquin, Lanctot, Schaul,
  Piot, Sendonaris, Dulac{-}Arnold, Osband, Agapiou, Leibo, and Gruslys}{Hester
  et~al\mbox{.}}{2017}]%
        {Hester17}
\bibfield{author}{\bibinfo{person}{Todd Hester}, \bibinfo{person}{Matej
  Vecerik}, \bibinfo{person}{Olivier Pietquin}, \bibinfo{person}{Marc Lanctot},
  \bibinfo{person}{Tom Schaul}, \bibinfo{person}{Bilal Piot},
  \bibinfo{person}{Andrew Sendonaris}, \bibinfo{person}{Gabriel
  Dulac{-}Arnold}, \bibinfo{person}{Ian Osband}, \bibinfo{person}{John
  Agapiou}, \bibinfo{person}{Joel~Z. Leibo}, {and} \bibinfo{person}{Audrunas
  Gruslys}.} \bibinfo{year}{2017}\natexlab{}.
\newblock \showarticletitle{Learning from Demonstrations for Real World
  Reinforcement Learning}.
\newblock \bibinfo{journal}{\emph{CoRR}}  \bibinfo{volume}{abs/1704.03732}
  (\bibinfo{year}{2017}).
\newblock
\showeprint[arxiv]{1704.03732}
\urldef\tempurl%
\url{http://arxiv.org/abs/1704.03732}
\showURL{%
\tempurl}


\bibitem[\protect\citeauthoryear{Ho and Ermon}{Ho and Ermon}{[n. d.]}]%
        {Ho2016}
\bibfield{author}{\bibinfo{person}{Jonathan Ho} {and} \bibinfo{person}{Stefano
  Ermon}.} \bibinfo{year}{[n. d.]}\natexlab{}.
\newblock \showarticletitle{Generative Adversarial Imitation Learning}.
\newblock  (\bibinfo{year}{[n. d.]}).
\newblock
\showeprint[arXiv]{cs.LG/1606.03476v1}


\bibitem[\protect\citeauthoryear{Idreos, Manegold, Kuno, and Graefe}{Idreos
  et~al\mbox{.}}{2011}]%
        {Idreos2011}
\bibfield{author}{\bibinfo{person}{Stratos Idreos}, \bibinfo{person}{Stefan
  Manegold}, \bibinfo{person}{Harumi Kuno}, {and} \bibinfo{person}{Goetz
  Graefe}.} \bibinfo{year}{2011}\natexlab{}.
\newblock \showarticletitle{Merging What's Cracked, Cracking What's Merged:
  Adaptive Indexing in Main-memory Column-stores}.
\newblock \bibinfo{journal}{\emph{Proc. VLDB Endow.}} \bibinfo{volume}{4},
  \bibinfo{number}{9} (\bibinfo{date}{June} \bibinfo{year}{2011}),
  \bibinfo{pages}{586--597}.
\newblock
\showISSN{2150-8097}
\urldef\tempurl%
\url{https://doi.org/10.14778/2002938.2002944}
\showDOI{\tempurl}


\bibitem[\protect\citeauthoryear{imdb.com}{imdb.com}{2018}]%
        {imdb}
\bibfield{author}{\bibinfo{person}{imdb.com}.} \bibinfo{year}{2018}\natexlab{}.
\newblock \bibinfo{title}{IMDb Datasets}.
\newblock \bibinfo{howpublished}{website}.   (\bibinfo{year}{2018}).
\newblock
\urldef\tempurl%
\url{https://www.imdb.com/interfaces/}
\showURL{%
\tempurl}


\bibitem[\protect\citeauthoryear{Kansky, Silver, M{\'e}ly, Eldawy,
  L{\'a}zaro-Gredilla, Lou, Dorfman, Sidor, Phoenix, and George}{Kansky
  et~al\mbox{.}}{[n. d.]}]%
        {Kansky2017}
\bibfield{author}{\bibinfo{person}{Ken Kansky}, \bibinfo{person}{Tom Silver},
  \bibinfo{person}{David~A. M{\'e}ly}, \bibinfo{person}{Mohamed Eldawy},
  \bibinfo{person}{Miguel L{\'a}zaro-Gredilla}, \bibinfo{person}{Xinghua Lou},
  \bibinfo{person}{Nimrod Dorfman}, \bibinfo{person}{Szymon Sidor},
  \bibinfo{person}{Scott Phoenix}, {and} \bibinfo{person}{Dileep George}.}
  \bibinfo{year}{[n. d.]}\natexlab{}.
\newblock \showarticletitle{Schema Networks: Zero-shot Transfer with a
  Generative Causal Model of Intuitive Physics}.
\newblock  (\bibinfo{year}{[n. d.]}).
\newblock
\showeprint[arXiv]{cs.AI/1706.04317v2}


\bibitem[\protect\citeauthoryear{Kingma and Ba}{Kingma and Ba}{2014}]%
        {KingmaBa2014}
\bibfield{author}{\bibinfo{person}{Diederik Kingma} {and}
  \bibinfo{person}{Jimmy Ba}.} \bibinfo{year}{2014}\natexlab{}.
\newblock \showarticletitle{Adam: A Method for Stochastic Optimization}.
\newblock  (\bibinfo{date}{Dec.} \bibinfo{year}{2014}).
\newblock
\showeprint{1412.6980}


\bibitem[\protect\citeauthoryear{Kraska, Beutel, Chi, Dean, and
  Polyzotis}{Kraska et~al\mbox{.}}{[n. d.]}]%
        {Kraska2017}
\bibfield{author}{\bibinfo{person}{Tim Kraska}, \bibinfo{person}{Alex Beutel},
  \bibinfo{person}{Ed~H. Chi}, \bibinfo{person}{Jeffrey Dean}, {and}
  \bibinfo{person}{Neoklis Polyzotis}.} \bibinfo{year}{[n. d.]}\natexlab{}.
\newblock \showarticletitle{The Case for Learned Index Structures}.
\newblock  (\bibinfo{year}{[n. d.]}).
\newblock
\showeprint[arXiv]{cs.DB/1712.01208v2}


\bibitem[\protect\citeauthoryear{Kulkarni, Bhagat, Fu, Kedigehalli, Kellogg,
  Mittal, Patel, Ramasamy, and Taneja}{Kulkarni et~al\mbox{.}}{2015}]%
        {Kulkarni2015}
\bibfield{author}{\bibinfo{person}{Sanjeev Kulkarni}, \bibinfo{person}{Nikunj
  Bhagat}, \bibinfo{person}{Maosong Fu}, \bibinfo{person}{Vikas Kedigehalli},
  \bibinfo{person}{Christopher Kellogg}, \bibinfo{person}{Sailesh Mittal},
  \bibinfo{person}{Jignesh~M Patel}, \bibinfo{person}{Karthik Ramasamy}, {and}
  \bibinfo{person}{Siddarth Taneja}.} \bibinfo{year}{2015}\natexlab{}.
\newblock \showarticletitle{Twitter heron: Stream processing at scale}. In
  \bibinfo{booktitle}{\emph{Proceedings of the 2015 ACM SIGMOD International
  Conference on Management of Data}}. ACM, \bibinfo{pages}{239--250}.
\newblock


\bibitem[\protect\citeauthoryear{Kumar and Miikkulainen}{Kumar and
  Miikkulainen}{1997}]%
        {KumarMiikkulainen1997}
\bibfield{author}{\bibinfo{person}{Shailesh Kumar} {and} \bibinfo{person}{Risto
  Miikkulainen}.} \bibinfo{year}{1997}\natexlab{}.
\newblock \showarticletitle{Dual reinforcement Q-routing: An on-line adaptive
  routing algorithm}. In \bibinfo{booktitle}{\emph{Artificial neural networks
  in engineering}}.
\newblock


\bibitem[\protect\citeauthoryear{Kumar and Miikkulainen}{Kumar and
  Miikkulainen}{1999}]%
        {KumarMiikkulainen1999}
\bibfield{author}{\bibinfo{person}{Shailesh Kumar} {and} \bibinfo{person}{Risto
  Miikkulainen}.} \bibinfo{year}{1999}\natexlab{}.
\newblock \showarticletitle{Confidence based dual reinforcement q-routing: An
  adaptive online network routing algorithm}. In
  \bibinfo{booktitle}{\emph{IJCAI}}, Vol.~\bibinfo{volume}{99}. Citeseer,
  \bibinfo{pages}{758--763}.
\newblock


\bibitem[\protect\citeauthoryear{Li}{Li}{2017}]%
        {li2017deep}
\bibfield{author}{\bibinfo{person}{Yuxi Li}.} \bibinfo{year}{2017}\natexlab{}.
\newblock \showarticletitle{Deep reinforcement learning: An overview}.
\newblock \bibinfo{journal}{\emph{arXiv preprint arXiv:1701.07274}}
  (\bibinfo{year}{2017}).
\newblock


\bibitem[\protect\citeauthoryear{Liang, Liaw, Nishihara, Moritz, Fox, Gonzalez,
  Goldberg, and Stoica}{Liang et~al\mbox{.}}{2017}]%
        {Liang2017}
\bibfield{author}{\bibinfo{person}{Eric Liang}, \bibinfo{person}{Richard Liaw},
  \bibinfo{person}{Robert Nishihara}, \bibinfo{person}{Philipp Moritz},
  \bibinfo{person}{Roy Fox}, \bibinfo{person}{Joseph Gonzalez},
  \bibinfo{person}{Ken Goldberg}, {and} \bibinfo{person}{Ion Stoica}.}
  \bibinfo{year}{2017}\natexlab{}.
\newblock \showarticletitle{Ray RLLib: A Composable and Scalable Reinforcement
  Learning Library}.
\newblock \bibinfo{journal}{\emph{arXiv preprint arXiv:1712.09381}}
  (\bibinfo{year}{2017}).
\newblock


\bibitem[\protect\citeauthoryear{Liu, Lin, Zhou, Nan, Liu, and Du}{Liu
  et~al\mbox{.}}{2018}]%
        {Liu2018}
\bibfield{author}{\bibinfo{person}{Sicong Liu}, \bibinfo{person}{Yingyan Lin},
  \bibinfo{person}{Zimu Zhou}, \bibinfo{person}{Kaiming Nan},
  \bibinfo{person}{Hui Liu}, {and} \bibinfo{person}{Junzhao Du}.}
  \bibinfo{year}{2018}\natexlab{}.
\newblock \showarticletitle{On-Demand Deep Model Compression for Mobile
  Devices: A Usage-Driven Model Selection Framework}.
\newblock  (\bibinfo{year}{2018}).
\newblock


\bibitem[\protect\citeauthoryear{Ma, Van~Aken, Hefny, Mezerhane, Pavlo, and
  Gordon}{Ma et~al\mbox{.}}{2018}]%
        {Ma2018}
\bibfield{author}{\bibinfo{person}{Lin Ma}, \bibinfo{person}{Dana Van~Aken},
  \bibinfo{person}{Ahmed Hefny}, \bibinfo{person}{Gustavo Mezerhane},
  \bibinfo{person}{Andrew Pavlo}, {and} \bibinfo{person}{Geoffrey~J. Gordon}.}
  \bibinfo{year}{2018}\natexlab{}.
\newblock \showarticletitle{Query-based Workload Forecasting for Self-Driving
  Database Management Systems}. In \bibinfo{booktitle}{\emph{Proceedings of the
  2018 International Conference on Management of Data}}
  \emph{(\bibinfo{series}{SIGMOD '18})}. \bibinfo{publisher}{ACM},
  \bibinfo{address}{New York, NY, USA}, \bibinfo{pages}{631--645}.
\newblock
\showISBNx{978-1-4503-4703-7}
\urldef\tempurl%
\url{https://doi.org/10.1145/3183713.3196908}
\showDOI{\tempurl}


\bibitem[\protect\citeauthoryear{Mania, Guy, and Recht}{Mania
  et~al\mbox{.}}{[n. d.]}]%
        {Mania2018}
\bibfield{author}{\bibinfo{person}{Horia Mania}, \bibinfo{person}{Aurelia Guy},
  {and} \bibinfo{person}{Benjamin Recht}.} \bibinfo{year}{[n. d.]}\natexlab{}.
\newblock \showarticletitle{Simple random search provides a competitive
  approach to reinforcement learning}.
\newblock  (\bibinfo{year}{[n. d.]}).
\newblock
\showeprint[arXiv]{cs.LG/1803.07055v1}


\bibitem[\protect\citeauthoryear{Mao, Alizadeh, Menache, and Kandula}{Mao
  et~al\mbox{.}}{2016}]%
        {Mao2016}
\bibfield{author}{\bibinfo{person}{Hongzi Mao}, \bibinfo{person}{Mohammad
  Alizadeh}, \bibinfo{person}{Ishai Menache}, {and} \bibinfo{person}{Srikanth
  Kandula}.} \bibinfo{year}{2016}\natexlab{}.
\newblock \showarticletitle{Resource Management with Deep Reinforcement
  Learning}. In \bibinfo{booktitle}{\emph{Proceedings of the 15th ACM Workshop
  on Hot Topics in Networks}}. ACM, \bibinfo{pages}{50--56}.
\newblock


\bibitem[\protect\citeauthoryear{Mao, Netravali, and Alizadeh}{Mao
  et~al\mbox{.}}{2017}]%
        {Mao2017}
\bibfield{author}{\bibinfo{person}{Hongzi Mao}, \bibinfo{person}{Ravi
  Netravali}, {and} \bibinfo{person}{Mohammad Alizadeh}.}
  \bibinfo{year}{2017}\natexlab{}.
\newblock \showarticletitle{Neural Adaptive Video Streaming with Pensieve}. In
  \bibinfo{booktitle}{\emph{Proceedings of the Conference of the ACM Special
  Interest Group on Data Communication}} \emph{(\bibinfo{series}{SIGCOMM
  '17})}. \bibinfo{publisher}{ACM}, \bibinfo{address}{New York, NY, USA},
  \bibinfo{pages}{197--210}.
\newblock
\urldef\tempurl%
\url{https://doi.org/10.1145/3098822.3098843}
\showDOI{\tempurl}


\bibitem[\protect\citeauthoryear{Microsoft}{Microsoft}{2018}]%
        {cosmosdb}
\bibfield{author}{\bibinfo{person}{Microsoft}.}
  \bibinfo{year}{2018}\natexlab{}.
\newblock \bibinfo{title}{CosmosDB - A globally distributed database for low
  latency and massively scalable applications, with native support for NoSQL}.
\newblock \bibinfo{howpublished}{website}.   (\bibinfo{year}{2018}).
\newblock
\urldef\tempurl%
\url{https://azure.microsoft.com/en-gb/services/cosmos-db/}
\showURL{%
\tempurl}


\bibitem[\protect\citeauthoryear{Mikolov, Chen, Corrado, and Dean}{Mikolov
  et~al\mbox{.}}{[n. d.]}]%
        {Mikolov2013}
\bibfield{author}{\bibinfo{person}{Tomas Mikolov}, \bibinfo{person}{Kai Chen},
  \bibinfo{person}{Greg Corrado}, {and} \bibinfo{person}{Jeffrey Dean}.}
  \bibinfo{year}{[n. d.]}\natexlab{}.
\newblock \showarticletitle{Efficient Estimation of Word Representations in
  Vector Space}.
\newblock  (\bibinfo{year}{[n. d.]}).
\newblock
\showeprint[arXiv]{cs.CL/1301.3781v3}


\bibitem[\protect\citeauthoryear{Mirhoseini, Goldie, Pham, Steiner, Le, and
  Dean}{Mirhoseini et~al\mbox{.}}{2018}]%
        {hierarchical2018}
\bibfield{author}{\bibinfo{person}{Azalia Mirhoseini}, \bibinfo{person}{Anna
  Goldie}, \bibinfo{person}{Hieu Pham}, \bibinfo{person}{Benoit Steiner},
  \bibinfo{person}{Quoc~V. Le}, {and} \bibinfo{person}{Jeff Dean}.}
  \bibinfo{year}{2018}\natexlab{}.
\newblock \showarticletitle{Hierarchical Planning for Device Placement}.
\newblock
\urldef\tempurl%
\url{https://openreview.net/pdf?id=Hkc-TeZ0W}
\showURL{%
\tempurl}


\bibitem[\protect\citeauthoryear{Mirhoseini, Pham, Le, Steiner, Larsen, Zhou,
  Kumar, Norouzi, Bengio, and Dean}{Mirhoseini et~al\mbox{.}}{2017}]%
        {mirhoseini2017device}
\bibfield{author}{\bibinfo{person}{Azalia Mirhoseini}, \bibinfo{person}{Hieu
  Pham}, \bibinfo{person}{Quoc~V Le}, \bibinfo{person}{Benoit Steiner},
  \bibinfo{person}{Rasmus Larsen}, \bibinfo{person}{Yuefeng Zhou},
  \bibinfo{person}{Naveen Kumar}, \bibinfo{person}{Mohammad Norouzi},
  \bibinfo{person}{Samy Bengio}, {and} \bibinfo{person}{Jeff Dean}.}
  \bibinfo{year}{2017}\natexlab{}.
\newblock \showarticletitle{Device Placement Optimization with Reinforcement
  Learning}.
\newblock \bibinfo{journal}{\emph{arXiv preprint arXiv:1706.04972}}
  (\bibinfo{year}{2017}).
\newblock


\bibitem[\protect\citeauthoryear{Mnih, Badia, Mirza, Graves, Lillicrap, Harley,
  Silver, and Kavukcuoglu}{Mnih et~al\mbox{.}}{2016}]%
        {MnihBadiaMirzaEtAl2016}
\bibfield{author}{\bibinfo{person}{Volodymyr Mnih},
  \bibinfo{person}{Adrià~Puigdomènech Badia}, \bibinfo{person}{Mehdi Mirza},
  \bibinfo{person}{Alex Graves}, \bibinfo{person}{Timothy~P. Lillicrap},
  \bibinfo{person}{Tim Harley}, \bibinfo{person}{David Silver}, {and}
  \bibinfo{person}{Koray Kavukcuoglu}.} \bibinfo{year}{2016}\natexlab{}.
\newblock \showarticletitle{Asynchronous Methods for Deep Reinforcement
  Learning}.
\newblock  (\bibinfo{date}{Feb.} \bibinfo{year}{2016}).
\newblock
\showeprint{1602.01783}


\bibitem[\protect\citeauthoryear{Mnih, Kavukcuoglu, Silver, Graves, Antonoglou,
  Wierstra, and Riedmiller}{Mnih et~al\mbox{.}}{2013}]%
        {MnihKavukcuogluSilverEtAl2013}
\bibfield{author}{\bibinfo{person}{Volodymyr Mnih}, \bibinfo{person}{Koray
  Kavukcuoglu}, \bibinfo{person}{David Silver}, \bibinfo{person}{Alex Graves},
  \bibinfo{person}{Ioannis Antonoglou}, \bibinfo{person}{Daan Wierstra}, {and}
  \bibinfo{person}{Martin Riedmiller}.} \bibinfo{year}{2013}\natexlab{}.
\newblock \showarticletitle{Playing Atari with Deep Reinforcement Learning}.
\newblock \bibinfo{journal}{\emph{arXiv preprint arXiv:1312.5602}}
  (\bibinfo{year}{2013}).
\newblock


\bibitem[\protect\citeauthoryear{Mnih, Kavukcuoglu, Silver, Rusu, Veness,
  Bellemare, Graves, Riedmiller, Fidjeland, Ostrovski, et~al\mbox{.}}{Mnih
  et~al\mbox{.}}{2015}]%
        {MnihDQN2015}
\bibfield{author}{\bibinfo{person}{Volodymyr Mnih}, \bibinfo{person}{Koray
  Kavukcuoglu}, \bibinfo{person}{David Silver}, \bibinfo{person}{Andrei~A
  Rusu}, \bibinfo{person}{Joel Veness}, \bibinfo{person}{Marc~G Bellemare},
  \bibinfo{person}{Alex Graves}, \bibinfo{person}{Martin Riedmiller},
  \bibinfo{person}{Andreas~K Fidjeland}, \bibinfo{person}{Georg Ostrovski},
  {et~al\mbox{.}}} \bibinfo{year}{2015}\natexlab{}.
\newblock \showarticletitle{Human-level control through deep reinforcement
  learning}.
\newblock \bibinfo{journal}{\emph{Nature}} \bibinfo{volume}{518},
  \bibinfo{number}{7540} (\bibinfo{year}{2015}), \bibinfo{pages}{529--533}.
\newblock


\bibitem[\protect\citeauthoryear{{ObjectLabs Corporation}}{{ObjectLabs
  Corporation}}{2018}]%
        {mlab2018}
\bibfield{author}{\bibinfo{person}{{ObjectLabs Corporation}}.}
  \bibinfo{year}{2018}\natexlab{}.
\newblock \bibinfo{title}{mLab cloud service}.
\newblock \bibinfo{howpublished}{website}.   (\bibinfo{year}{2018}).
\newblock
\urldef\tempurl%
\url{https://mlab.com}
\showURL{%
\tempurl}


\bibitem[\protect\citeauthoryear{OpenAI}{OpenAI}{2018}]%
        {openaidota}
\bibfield{author}{\bibinfo{person}{OpenAI}.} \bibinfo{year}{2018}\natexlab{}.
\newblock \bibinfo{title}{OpenAI Five DOTA}.
\newblock \bibinfo{howpublished}{website}.   (\bibinfo{date}{June}
  \bibinfo{year}{2018}).
\newblock
\urldef\tempurl%
\url{https://blog.openai.com/openai-five/}
\showURL{%
\tempurl}


\bibitem[\protect\citeauthoryear{Pavlo, Angulo, Arulraj, Lin, Lin, Ma, Menon,
  Mowry, Perron, Quah, et~al\mbox{.}}{Pavlo et~al\mbox{.}}{2017}]%
        {Pavlo2017}
\bibfield{author}{\bibinfo{person}{Andrew Pavlo}, \bibinfo{person}{Gustavo
  Angulo}, \bibinfo{person}{Joy Arulraj}, \bibinfo{person}{Haibin Lin},
  \bibinfo{person}{Jiexi Lin}, \bibinfo{person}{Lin Ma},
  \bibinfo{person}{Prashanth Menon}, \bibinfo{person}{Todd~C Mowry},
  \bibinfo{person}{Matthew Perron}, \bibinfo{person}{Ian Quah},
  {et~al\mbox{.}}} \bibinfo{year}{2017}\natexlab{}.
\newblock \showarticletitle{Self-Driving Database Management Systems.}. In
  \bibinfo{booktitle}{\emph{CIDR}}.
\newblock


\bibitem[\protect\citeauthoryear{{Peter Bailis et al.}}{{Peter Bailis et
  al.}}{2018}]%
        {bailis2018learning}
\bibfield{author}{\bibinfo{person}{{Peter Bailis et al.}}}
  \bibinfo{year}{2018}\natexlab{}.
\newblock \bibinfo{title}{Don't Throw Out Your Algorithms Book Just Yet:
  Classical Data Structures That Can Outperform Learned Indexes}.
\newblock \bibinfo{howpublished}{website}.   (\bibinfo{date}{Jan.}
  \bibinfo{year}{2018}).
\newblock
\urldef\tempurl%
\url{https://dawn.cs.stanford.edu/2018/01/11/index-baselines/}
\showURL{%
\tempurl}


\bibitem[\protect\citeauthoryear{Petraki, Idreos, and Manegold}{Petraki
  et~al\mbox{.}}{2015}]%
        {Petraki2015}
\bibfield{author}{\bibinfo{person}{Eleni Petraki}, \bibinfo{person}{Stratos
  Idreos}, {and} \bibinfo{person}{Stefan Manegold}.}
  \bibinfo{year}{2015}\natexlab{}.
\newblock \showarticletitle{Holistic Indexing in Main-memory Column-stores}. In
  \bibinfo{booktitle}{\emph{Proceedings of the 2015 ACM SIGMOD International
  Conference on Management of Data}} \emph{(\bibinfo{series}{SIGMOD '15})}.
  \bibinfo{publisher}{ACM}, \bibinfo{address}{New York, NY, USA},
  \bibinfo{pages}{1153--1166}.
\newblock
\showISBNx{978-1-4503-2758-9}
\urldef\tempurl%
\url{https://doi.org/10.1145/2723372.2723719}
\showDOI{\tempurl}


\bibitem[\protect\citeauthoryear{Piot, Geist, and Pietquin}{Piot
  et~al\mbox{.}}{2014}]%
        {Piot2014}
\bibfield{author}{\bibinfo{person}{Bilal Piot}, \bibinfo{person}{Matthieu
  Geist}, {and} \bibinfo{person}{Olivier Pietquin}.}
  \bibinfo{year}{2014}\natexlab{}.
\newblock \showarticletitle{Boosted Bellman residual minimization handling
  expert demonstrations}. In \bibinfo{booktitle}{\emph{Joint European
  Conference on Machine Learning and Knowledge Discovery in Databases}}.
  Springer, \bibinfo{pages}{549--564}.
\newblock


\bibitem[\protect\citeauthoryear{Qader, Cheng, and Hristidis}{Qader
  et~al\mbox{.}}{2018}]%
        {Qader2018}
\bibfield{author}{\bibinfo{person}{Mohiuddin~Abdul Qader},
  \bibinfo{person}{Shiwen Cheng}, {and} \bibinfo{person}{Vagelis Hristidis}.}
  \bibinfo{year}{2018}\natexlab{}.
\newblock \showarticletitle{A Comparative Study of Secondary Indexing
  Techniques in LSM-based NoSQL Databases}. In
  \bibinfo{booktitle}{\emph{Proceedings of the 2018 International Conference on
  Management of Data}} \emph{(\bibinfo{series}{SIGMOD '18})}.
  \bibinfo{publisher}{ACM}, \bibinfo{address}{New York, NY, USA},
  \bibinfo{pages}{551--566}.
\newblock
\showISBNx{978-1-4503-4703-7}
\urldef\tempurl%
\url{https://doi.org/10.1145/3183713.3196900}
\showDOI{\tempurl}


\bibitem[\protect\citeauthoryear{Ratner, Bach, Ehrenberg, Fries, Wu, and
  R{\'e}}{Ratner et~al\mbox{.}}{2017}]%
        {Ratner2017}
\bibfield{author}{\bibinfo{person}{Alexander Ratner},
  \bibinfo{person}{Stephen~H Bach}, \bibinfo{person}{Henry Ehrenberg},
  \bibinfo{person}{Jason Fries}, \bibinfo{person}{Sen Wu}, {and}
  \bibinfo{person}{Christopher R{\'e}}.} \bibinfo{year}{2017}\natexlab{}.
\newblock \showarticletitle{Snorkel: Rapid training data creation with weak
  supervision}.
\newblock \bibinfo{journal}{\emph{Proceedings of the VLDB Endowment}}
  \bibinfo{volume}{11}, \bibinfo{number}{3} (\bibinfo{year}{2017}),
  \bibinfo{pages}{269--282}.
\newblock


\bibitem[\protect\citeauthoryear{Ross, Gordon, and Bagnell}{Ross
  et~al\mbox{.}}{[n. d.]}]%
        {Ross2010}
\bibfield{author}{\bibinfo{person}{Stephane Ross}, \bibinfo{person}{Geoffrey~J.
  Gordon}, {and} \bibinfo{person}{J.~Andrew Bagnell}.} \bibinfo{year}{[n.
  d.]}\natexlab{}.
\newblock \showarticletitle{A Reduction of Imitation Learning and Structured
  Prediction to No-Regret Online Learning}.
\newblock  (\bibinfo{year}{[n. d.]}).
\newblock
\showeprint[arXiv]{cs.LG/1011.0686v3}


\bibitem[\protect\citeauthoryear{Schnaitter, Abiteboul, Milo, and
  Polyzotis}{Schnaitter et~al\mbox{.}}{2006}]%
        {Schnaitter2006}
\bibfield{author}{\bibinfo{person}{Karl Schnaitter}, \bibinfo{person}{Serge
  Abiteboul}, \bibinfo{person}{Tova Milo}, {and} \bibinfo{person}{Neoklis
  Polyzotis}.} \bibinfo{year}{2006}\natexlab{}.
\newblock \showarticletitle{COLT: Continuous On-line Tuning}. In
  \bibinfo{booktitle}{\emph{Proceedings of the 2006 ACM SIGMOD International
  Conference on Management of Data}} \emph{(\bibinfo{series}{SIGMOD '06})}.
  \bibinfo{publisher}{ACM}, \bibinfo{address}{New York, NY, USA},
  \bibinfo{pages}{793--795}.
\newblock
\showISBNx{1-59593-434-0}
\urldef\tempurl%
\url{https://doi.org/10.1145/1142473.1142592}
\showDOI{\tempurl}


\bibitem[\protect\citeauthoryear{Schulman, Levine, Abbeel, Jordan, and
  Moritz}{Schulman et~al\mbox{.}}{2015}]%
        {schulman2015trust}
\bibfield{author}{\bibinfo{person}{John Schulman}, \bibinfo{person}{Sergey
  Levine}, \bibinfo{person}{Pieter Abbeel}, \bibinfo{person}{Michael Jordan},
  {and} \bibinfo{person}{Philipp Moritz}.} \bibinfo{year}{2015}\natexlab{}.
\newblock \showarticletitle{Trust region policy optimization}. In
  \bibinfo{booktitle}{\emph{Proceedings of the 32nd International Conference on
  Machine Learning (ICML-15)}}. \bibinfo{pages}{1889--1897}.
\newblock


\bibitem[\protect\citeauthoryear{Schulman, Wolski, Dhariwal, Radford, and
  Klimov}{Schulman et~al\mbox{.}}{2017}]%
        {schulman2017proximal}
\bibfield{author}{\bibinfo{person}{John Schulman}, \bibinfo{person}{Filip
  Wolski}, \bibinfo{person}{Prafulla Dhariwal}, \bibinfo{person}{Alec Radford},
  {and} \bibinfo{person}{Oleg Klimov}.} \bibinfo{year}{2017}\natexlab{}.
\newblock \showarticletitle{Proximal Policy Optimization Algorithms}.
\newblock \bibinfo{journal}{\emph{arXiv preprint arXiv:1707.06347}}
  (\bibinfo{year}{2017}).
\newblock


\bibitem[\protect\citeauthoryear{Seide and Agarwal}{Seide and Agarwal}{2016}]%
        {seide2016cntk}
\bibfield{author}{\bibinfo{person}{Frank Seide} {and} \bibinfo{person}{Amit
  Agarwal}.} \bibinfo{year}{2016}\natexlab{}.
\newblock \showarticletitle{CNTK: Microsoft's Open-Source Deep-Learning
  Toolkit}. In \bibinfo{booktitle}{\emph{Proceedings of the 22nd ACM SIGKDD
  International Conference on Knowledge Discovery and Data Mining}}. ACM,
  \bibinfo{pages}{2135--2135}.
\newblock


\bibitem[\protect\citeauthoryear{Sharma, Schuhknecht, and Dittrich}{Sharma
  et~al\mbox{.}}{[n. d.]}]%
        {Sharma2018}
\bibfield{author}{\bibinfo{person}{Ankur Sharma}, \bibinfo{person}{Felix~Martin
  Schuhknecht}, {and} \bibinfo{person}{Jens Dittrich}.} \bibinfo{year}{[n.
  d.]}\natexlab{}.
\newblock \showarticletitle{The Case for Automatic Database Administration
  using Deep Reinforcement Learning}.
\newblock  (\bibinfo{year}{[n. d.]}).
\newblock
\showeprint[arXiv]{cs.DB/http://arxiv.org/abs/1801.05643v1}


\bibitem[\protect\citeauthoryear{Sidor and Schulman}{Sidor and
  Schulman}{2017}]%
        {openaibaselines}
\bibfield{author}{\bibinfo{person}{Szymon Sidor} {and} \bibinfo{person}{John
  Schulman}.} \bibinfo{year}{2017}\natexlab{}.
\newblock \bibinfo{title}{OpenAI Baselines}.
\newblock \bibinfo{howpublished}{website}.   (\bibinfo{year}{2017}).
\newblock
\urldef\tempurl%
\url{https://blog.openai.com/openai-baselines-dqn/}
\showURL{%
\tempurl}


\bibitem[\protect\citeauthoryear{Silver, Huang, Maddison, Guez, Sifre, Van
  Den~Driessche, Schrittwieser, Antonoglou, Panneershelvam, Lanctot,
  et~al\mbox{.}}{Silver et~al\mbox{.}}{2016}]%
        {SilverHuangMaddisonEtAl2016}
\bibfield{author}{\bibinfo{person}{David Silver}, \bibinfo{person}{Aja Huang},
  \bibinfo{person}{Chris~J Maddison}, \bibinfo{person}{Arthur Guez},
  \bibinfo{person}{Laurent Sifre}, \bibinfo{person}{George Van Den~Driessche},
  \bibinfo{person}{Julian Schrittwieser}, \bibinfo{person}{Ioannis Antonoglou},
  \bibinfo{person}{Veda Panneershelvam}, \bibinfo{person}{Marc Lanctot},
  {et~al\mbox{.}}} \bibinfo{year}{2016}\natexlab{}.
\newblock \showarticletitle{Mastering the game of Go with deep neural networks
  and tree search}.
\newblock \bibinfo{journal}{\emph{Nature}} \bibinfo{volume}{529},
  \bibinfo{number}{7587} (\bibinfo{year}{2016}), \bibinfo{pages}{484--489}.
\newblock


\bibitem[\protect\citeauthoryear{Sutton and Barto}{Sutton and Barto}{1998}]%
        {SuttonBarto1998}
\bibfield{author}{\bibinfo{person}{Richard~S Sutton} {and}
  \bibinfo{person}{Andrew~G Barto}.} \bibinfo{year}{1998}\natexlab{}.
\newblock \bibinfo{booktitle}{\emph{Reinforcement learning: An introduction}}.
  Vol.~\bibinfo{volume}{1}.
\newblock \bibinfo{publisher}{MIT press Cambridge}.
\newblock


\bibitem[\protect\citeauthoryear{Tesauro, Das, Chan, Kephart, Levine, Rawson,
  and Lefurgy}{Tesauro et~al\mbox{.}}{2007}]%
        {TesauroDasChanEtAl2007}
\bibfield{author}{\bibinfo{person}{Gerald Tesauro}, \bibinfo{person}{Rajarshi
  Das}, \bibinfo{person}{Hoi Chan}, \bibinfo{person}{Jeffrey Kephart},
  \bibinfo{person}{David Levine}, \bibinfo{person}{Freeman Rawson}, {and}
  \bibinfo{person}{Charles Lefurgy}.} \bibinfo{year}{2007}\natexlab{}.
\newblock \showarticletitle{Managing power consumption and performance of
  computing systems using reinforcement learning}. In
  \bibinfo{booktitle}{\emph{Advances in Neural Information Processing
  Systems}}. \bibinfo{pages}{1497--1504}.
\newblock


\bibitem[\protect\citeauthoryear{Tesauro, Jong, Das, and Bennani}{Tesauro
  et~al\mbox{.}}{2006}]%
        {TesauroJongDasEtAl2006}
\bibfield{author}{\bibinfo{person}{G. Tesauro}, \bibinfo{person}{N.~K. Jong},
  \bibinfo{person}{R. Das}, {and} \bibinfo{person}{M.~N. Bennani}.}
  \bibinfo{year}{2006}\natexlab{}.
\newblock \showarticletitle{A Hybrid Reinforcement Learning Approach to
  Autonomic Resource Allocation}. In \bibinfo{booktitle}{\emph{Proceedings of
  the 2006 IEEE International Conference on Autonomic Computing}}
  \emph{(\bibinfo{series}{ICAC '06})}. \bibinfo{publisher}{IEEE Computer
  Society}, \bibinfo{address}{Washington, DC, USA}, \bibinfo{pages}{65--73}.
\newblock
\showISBNx{1-4244-0175-5}
\urldef\tempurl%
\url{https://doi.org/10.1109/ICAC.2006.1662383}
\showDOI{\tempurl}


\bibitem[\protect\citeauthoryear{{The Apache Software Foundation}}{{The Apache
  Software Foundation}}{2018}]%
        {storm}
\bibfield{author}{\bibinfo{person}{{The Apache Software Foundation}}.}
  \bibinfo{year}{2018}\natexlab{}.
\newblock \bibinfo{title}{Apache {S}torm}.
\newblock \bibinfo{howpublished}{{https://storm.apache.org/}}.
  (\bibinfo{year}{2018}).
\newblock


\bibitem[\protect\citeauthoryear{Tobin, Zaremba, and Abbeel}{Tobin
  et~al\mbox{.}}{2017}]%
        {Tobin2017}
\bibfield{author}{\bibinfo{person}{Joshua Tobin}, \bibinfo{person}{Wojciech
  Zaremba}, {and} \bibinfo{person}{Pieter Abbeel}.}
  \bibinfo{year}{2017}\natexlab{}.
\newblock \showarticletitle{Domain Randomization and Generative Models for
  Robotic Grasping}.
\newblock \bibinfo{journal}{\emph{arXiv preprint arXiv:1710.06425}}
  (\bibinfo{year}{2017}).
\newblock


\bibitem[\protect\citeauthoryear{Valadarsky, Schapira, Shahaf, and
  Tamar}{Valadarsky et~al\mbox{.}}{2017}]%
        {Valadarsky2017a}
\bibfield{author}{\bibinfo{person}{Asaf Valadarsky}, \bibinfo{person}{Michael
  Schapira}, \bibinfo{person}{Dafna Shahaf}, {and} \bibinfo{person}{Aviv
  Tamar}.} \bibinfo{year}{2017}\natexlab{}.
\newblock \showarticletitle{A Machine Learning Approach to Routing}.
\newblock \bibinfo{journal}{\emph{arXiv preprint arXiv:1708.03074}}
  (\bibinfo{year}{2017}).
\newblock


\bibitem[\protect\citeauthoryear{Van~Aken, Pavlo, Gordon, and Zhang}{Van~Aken
  et~al\mbox{.}}{2017}]%
        {VanAken2017}
\bibfield{author}{\bibinfo{person}{Dana Van~Aken}, \bibinfo{person}{Andrew
  Pavlo}, \bibinfo{person}{Geoffrey~J Gordon}, {and} \bibinfo{person}{Bohan
  Zhang}.} \bibinfo{year}{2017}\natexlab{}.
\newblock \showarticletitle{Automatic Database Management System Tuning Through
  Large-scale Machine Learning}. In \bibinfo{booktitle}{\emph{Proceedings of
  the 2017 ACM International Conference on Management of Data}}. ACM,
  \bibinfo{pages}{1009--1024}.
\newblock


\bibitem[\protect\citeauthoryear{{van Hasselt}, {Guez}, and {Silver}}{{van
  Hasselt} et~al\mbox{.}}{2015}]%
        {double_dqn}
\bibfield{author}{\bibinfo{person}{H. {van Hasselt}}, \bibinfo{person}{A.
  {Guez}}, {and} \bibinfo{person}{D. {Silver}}.}
  \bibinfo{year}{2015}\natexlab{}.
\newblock \showarticletitle{{Deep Reinforcement Learning with Double
  Q-learning}}.
\newblock \bibinfo{journal}{\emph{ArXiv e-prints}} (\bibinfo{date}{Sept.}
  \bibinfo{year}{2015}).
\newblock
\showeprint[arxiv]{cs.LG/1509.06461}


\bibitem[\protect\citeauthoryear{Wang, Merel, Reed, Wayne, de~Freitas, and
  Heess}{Wang et~al\mbox{.}}{[n. d.]}]%
        {Wang2017}
\bibfield{author}{\bibinfo{person}{Ziyu Wang}, \bibinfo{person}{Josh Merel},
  \bibinfo{person}{Scott Reed}, \bibinfo{person}{Greg Wayne},
  \bibinfo{person}{Nando de Freitas}, {and} \bibinfo{person}{Nicolas Heess}.}
  \bibinfo{year}{[n. d.]}\natexlab{}.
\newblock \showarticletitle{Robust Imitation of Diverse Behaviors}.
\newblock  (\bibinfo{year}{[n. d.]}).
\newblock
\showeprint[arXiv]{cs.LG/1707.02747v2}


\bibitem[\protect\citeauthoryear{Williams}{Williams}{1992}]%
        {Williams1992}
\bibfield{author}{\bibinfo{person}{Ronald~J Williams}.}
  \bibinfo{year}{1992}\natexlab{}.
\newblock \showarticletitle{Simple statistical gradient-following algorithms
  for connectionist reinforcement learning}.
\newblock \bibinfo{journal}{\emph{Machine learning}} \bibinfo{volume}{8},
  \bibinfo{number}{3-4} (\bibinfo{year}{1992}), \bibinfo{pages}{229--256}.
\newblock


\end{thebibliography}


\appendix
\section{Synthetic client}
We proceed to describe the synthetic data layout and query generation procedure used in the scalability and generalization experiments. Synthetic documents each contained 15 attributes with 6 strings, 6 integers of different ranges, 2 date fields, and 1 string array for full text. 

Synthetic were generated by sampling between 1 and 3 attributes. For each attribute, a sub-expression was generated by sampling both a comparison operator and a value for the attribute, e.g.:
\begin{alltt}
sub_expr := \{\$gt: \{"attribute": "value"\}\}
\end{alltt}
Sub-expressions were then concatenated by uniformly sampling a logical operator, e.g.:
\begin{alltt}
expr := \{\$or: [sub_expr, sub_expr, sub_expr]\}
\end{alltt}
Finally, we sampled an aggregation operator from a discrete distribution where a limit without sort or count had 0.1 probability and sort/count 0.45 each, as sort and counts caused more difficult indexing decisions. If a sort aggregation was sampled, sorting was requested for each attribute in the query with 0.5 probability. 
Below is an example of a resulting query:
\begin{alltt}
Q := find(\{'$or': [\{'f2': \{'$eq': 'centimeter'\}\},
\{'f10': \{'$gte': \{'$date': 1394135731965\}\}\}]\})
.limit(10)
\end{alltt}
In summary, our synthetic query client enables users to generate query shapes of varying difficulty and size to investigate indexing behavior.
\end{document}